\documentclass[11pt]{article}

\usepackage[preprint]{acl}

\usepackage{times}
\usepackage{latexsym}

\usepackage[T1]{fontenc}

\usepackage[utf8]{inputenc}

\usepackage{microtype}

\usepackage{inconsolata}

\usepackage{graphicx}

\usepackage{enumitem}
\usepackage{booktabs}
\usepackage{multirow}
\usepackage{amsmath}
\usepackage{amssymb}
\usepackage[table,dvipsnames]{xcolor}
\usepackage{xspace}
\usepackage{pifont}
\usepackage[ruled,vlined,linesnumbered]{algorithm2e}
\usepackage{cleveref}
\usepackage{makecell}
\usepackage{url}

\colorlet{bestgreen}{SeaGreen!50}
\colorlet{secondgreen}{SeaGreen!30}

\usepackage{fontawesome5}

\hypersetup{
    citecolor=cyan,
    filecolor=magenta,      
    urlcolor=magenta,
    }

\newcommand{\nus}{\raisebox{-0.25ex}{\includegraphics[height=0.85em]{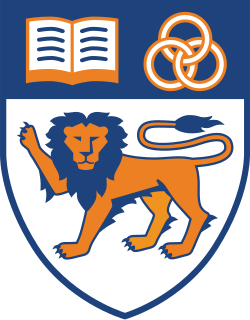}}}
\newcommand{\pku}{\raisebox{-0.25ex}{\includegraphics[height=0.85em]{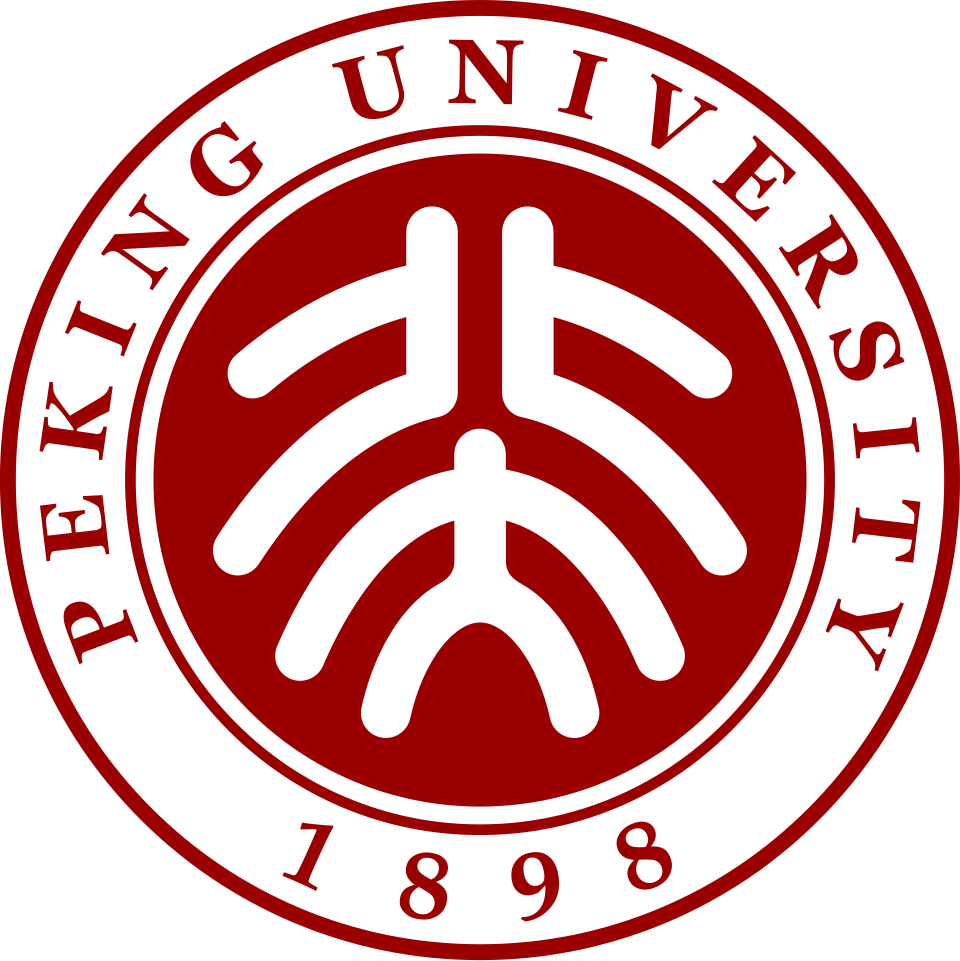}}}
\newcommand{\fdu}{\raisebox{-0.25ex}{\includegraphics[height=0.85em]{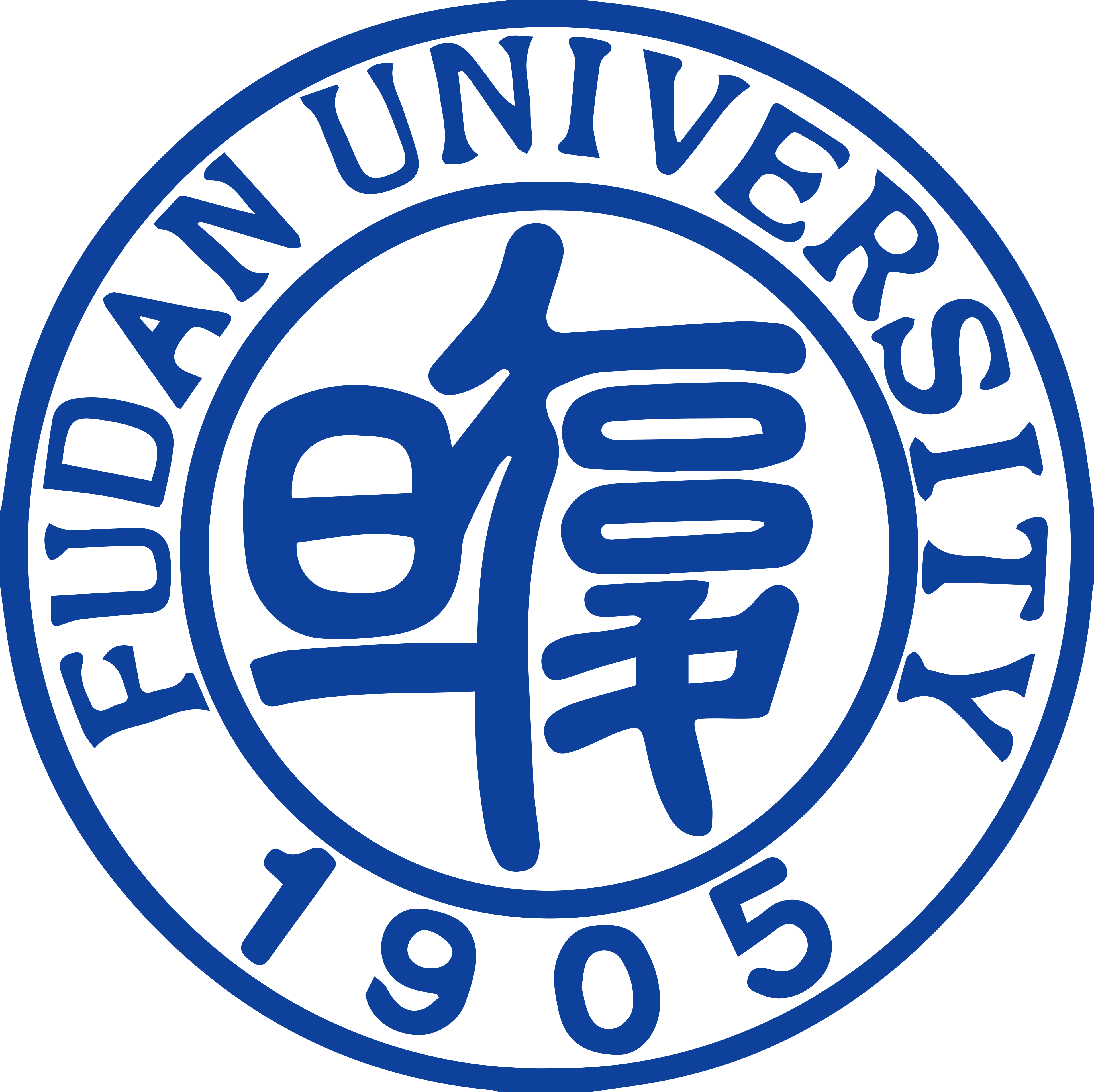}}}
\newcommand{\byte}{\raisebox{-0.25ex}{\includegraphics[height=0.85em]{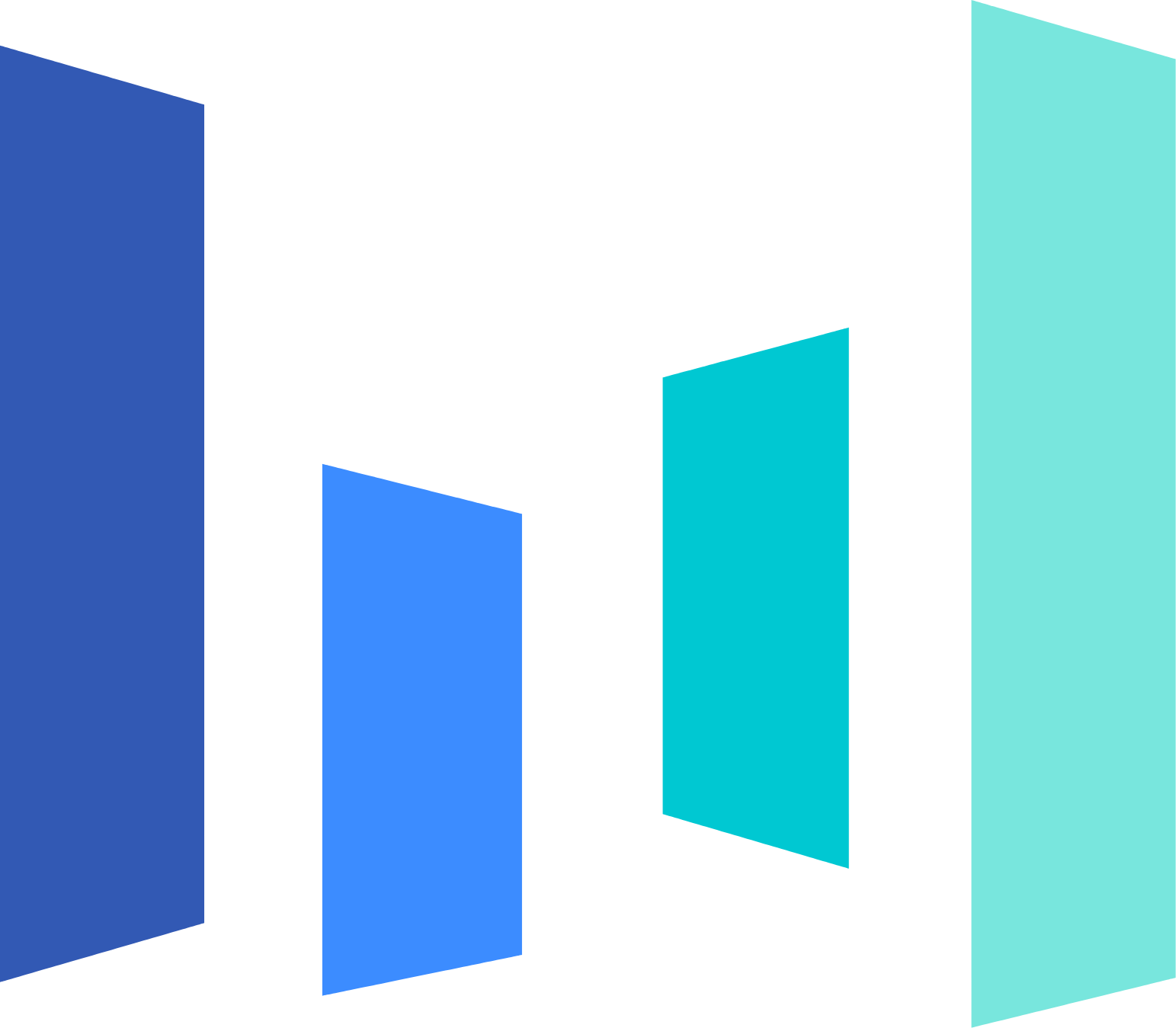}}}

\newcommand{\githubicon}{\raisebox{-1.5pt}{\includegraphics[height=1.05em]{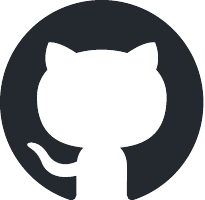}}}
\newcommand{\hficon}{\raisebox{-1.5pt}{\includegraphics[height=1.05em]{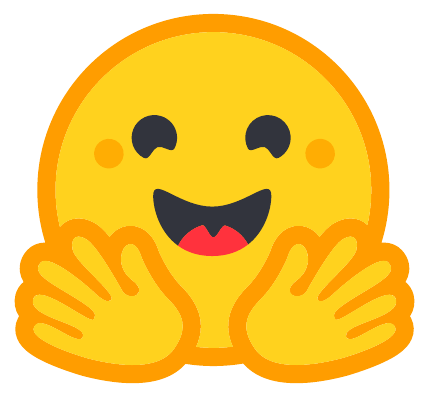}}}

\newcommand{\ghlink}{https://github.com/bingreeky/opd-evolver}
\newcommand{\hflink}{https://huggingface.co/greeky/OPDEvolver}
\newcommand{\projectlinks}{%
  \begin{tabular}{cc}
  \href{\ghlink}{\githubicon~GitHub} &
  \href{\hflink}{\hficon~Hugging Face} \\
  \end{tabular}
}

\makeatletter
\def\@fnsymbol#1{%
   \ifcase#1\or
   \TextOrMath \textdagger \dagger\or
   \TextOrMath \textdaggerdbl \ddagger \or
   \TextOrMath \textsection  \mathsection\or
   \TextOrMath \textparagraph \mathparagraph\or
   \TextOrMath \textbardbl \|\or
   \TextOrMath {\color{black}\textdagger\textdagger}{\dagger\dagger}\or
   \TextOrMath {\textdaggerdbl\textdaggerdbl}{\ddagger\ddagger}\else
   \@ctrerr \fi
}
\makeatother

\newcommand{\ourmethod}{\texttt{OPD-Evolver}\xspace}
\newcommand{\llmname}[1]{\textsc{#1}}
\newcommand{\algtt}[1]{\textcolor{cyan}{\texttt{#1}}}

\title{\ourmethod: Cultivating Holistic Agent Evolver via On-Policy Distillation}

\author{
Guibin Zhang{$^\dagger$}\quad
Xun Xu{$^\dagger$}\quad
Yanwei Yue\quad
Zikun Su\quad
Wangchunshu Zhou\\ 
\textbf{Xiaobin Hu}{$^\ddagger$}\quad
\textbf{Shuicheng Yan}{$^\ddagger$}\\
\vspace{0.2em}
\quad {$^\ddagger$} Corresponding Authors\; {$^\dagger$} Equal Contribution\\
 \nus~{LV-NUS Lab}\; \fdu~FDU\; \pku~PKU\; \byte~Bytedance Inc.\\
\projectlinks
}

\begin{document}
\maketitle
\begin{abstract}
Memory has become a standard substrate for self-evolving agents, yet retaining experience is not the same as learning how to evolve through it.
Existing memory agents can store trajectories, retrieve reflections, or accumulate skills, but often lack the holistic competence to select useful experience, act on it, write reusable knowledge, and maintain a growing repository.
We introduce \ourmethod, a slow-fast co-evolution framework that cultivates such an agent evolver through on-policy self-distillation.
In the fast loop, \ourmethod interacts with a four-level memory hierarchy to read, use, write, and maintain experience for rapid test-time evolution.
In the slow loop, outcome-calibrated memory attribution and privileged hindsight distill these four abilities into the deployable policy.
Across multi-domain benchmarks, \ourmethod surpasses memory systems such as ReasoningBank by up to $11.5\%$, and training-based methods such as Skill0 by $\sim5.8\%$.
Further analysis shows that \ourmethod internalizes high-value experience and memory management, enabling \ourmethod-9B to challenge giant counterparts such as \llmname{Qwen3.5-397B-A17B} and \llmname{Step-3.5-Flash}, pointing beyond memory-augmented agents toward genuinely qualified agent evolvers.
\end{abstract}

\title{\ourmethod: Cultivating a Holistic Agent Evolver via On-Policy Distillation}

\section{Introduction}
\label{sec:introduction}

\emph{What defines a self-evolving agent?}
A natural answer is memory, since preserving past trajectories, retrieving prior lessons, and reusing accumulated skills seem to provide the material basis for improvement over time~\citep{hu2025memoryageaiagentssurvey,fang2025comprehensivesurveyselfevolvingai}.
This intuition has made memory an indispensable component of modern agentic foundation models~\citep{kimiteam2025kimik2openagentic,minimaxMiniMaxM27} and agent systems~\citep{pan2026naturallanguageagentharnesses,zhou2026externalizationllmagentsunified}, especially in interactive environments where failures reveal latent constraints, successful rollouts expose reusable strategies, and rewards ground behavior in environmental feedback~\citep{wu2024streambenchbenchmarkingcontinuousimprovement,zheng2025lifelongagentbenchevaluatingllmagents,he2026memoryarenabenchmarkingagentmemory}.
Yet memory is only the substrate of self-evolution, not its definition.
Despite the proliferation of memory systems~\citep{zhang2025memevolvemetaevolutionagentmemory,ouyang2025reasoningbankscalingagentselfevolving}, skill libraries~\citep{zheng2025skillweaverwebagentsselfimprove,skill0,skillrl}, reflection mechanisms~\citep{liang2024self}, and self-improvement pipelines~\citep{evolver,evo-memory}, the notion of ``self-evolution'' remains under-specified: many agents can retain experience or expose it to the prompt, while far fewer can convert it into sustained behavioral improvement.
In this work, we define an \emph{agent evolver} as an agent that systematically transforms interaction history and feedback into persistent improvements in future behavior.

Existing work has approached this objective from diverse perspectives.
Memory-augmented agents store trajectories, reflections, tips, or lessons and inject them into later prompts~\citep{reflexion,expel,mem0}.
Skill-augmented agents, which we regard as a structured instantiation of memory, distill experience into reusable strategies, tools, or procedures~\citep{skillrl,memskill,skill0,shi2026skill1unifiedevolutionskillaugmented}.
Other methods more directly parameterize experience by training on collected trajectories through supervised fine-tuning (SFT), reinforcement learning (RL), or on-policy distillation~\citep{liu2026agenticcriticaltraining,zhang2025agentrlscalingagenticreinforcement,xi2025agentgymrltrainingllmagents}.
Despite this progress, most methods optimize only one fragment of the evolution process, such as retrieving experience, using it in context, distilling it into parameters, or engineering the memory architecture.
Two challenges remain central: task rewards provide relatively direct supervision for execution, but not for memory selection, writing, or long-term management; and how to train these coupled abilities within one policy without making them interfere remains underexplored.
As a result, existing agents may improve within a particular setting, yet still fall short of becoming holistic agent evolvers.

We characterize a qualified agent evolver through four coupled capabilities:
\ding{182} \textbf{experience selection} identifies useful memories from a growing and noisy repository; \ding{183} \textbf{experience-grounded execution} converts selected experience into effective multi-turn actions; \ding{184} \textbf{experience writing} extracts reusable knowledge from new trajectories and feedback; and \ding{185} \textbf{experience management} scores, consolidates, updates, and retires memories over time.
These competencies \textit{cannot be safely separated}: weak selection amplifies retrieval noise, weak execution leaves the agent dependent on prompt-time guidance, weak writing pollutes future memory, and weak management causes long-term degradation~\citep{zhang2025memevolvemetaevolutionagentmemory}.
This motivates our central research question:
\begin{center}
\fbox{
\begin{minipage}{0.92\linewidth}
\emph{How can we train an agent to acquire the holistic competence of evolving through experience, so as to become a qualified agent evolver?}
\end{minipage}
}
\end{center}

To this end, we introduce \ourmethod, a slow-fast co-evolution framework where the fast loop lets the agent evolve through online interaction with environments and memory, while the slow loop distills these interaction-derived behaviors into intrinsic evolver capabilities via on-policy self-distillation.
In the \ding{168} \textbf{fast evolution loop}, the agent operates over a four-level memory substrate of \textit{trajectories}, \textit{tips}, \textit{skills}, and \textit{tools}: it selects task-relevant memories before execution, acts with them, then writes and periodically maintains memories from trajectories, rewards, and feedback.
Across this stream, \ourmethod estimates an \emph{outcome-calibrated memory attribution} signal, converting delayed task outcomes into supervision for what should be selected, used, written, and preserved.

The \ding{171} \textbf{slow evolution loop} then turns this calibrated stream into on-policy self-distillation targets for the same policy.
The privileged teacher observes attribution-enriched evidence unavailable at deployment, including candidate memory value, trajectory snippets, future utility of written memories, and repository-level diagnostics.
Distilling the student's own rollout states against this hindsight supervision jointly shapes selection, execution, writing, and management, enabling \ourmethod to acquire transferable lifecycle-level competence for self-improvement.
Our contributions can be summarized as:

\vspace{-0.5em}
\begin{itemize}[leftmargin=0.2em,itemsep=-0.1em]
    \item[\ding{227}] \textbf{Objective Formulation.}
    We define the qualified agent evolver as an agent with four coupled competencies: experience selection, experience-grounded execution, experience writing, and experience management.

    \item[\ding{227}] \textbf{Practical Solution.}
    We propose \ourmethod, a fast-slow on-policy self-distillation framework that converts task outcomes and privileged hindsight into supervision for selection, execution, writing, and memory management.

    \item[\ding{227}] \textbf{Experimental Validation.}
    Evaluations on four self-evolving benchmarks show that \ourmethod-4B/9B surpass contemporary memory systems and remain competitive with giant counterparts such as \llmname{Qwen3.5-397B} and \llmname{Step-3.5-196B}.

\end{itemize}

\begin{figure*}[!t]
    \centering
    \includegraphics[width=\linewidth]{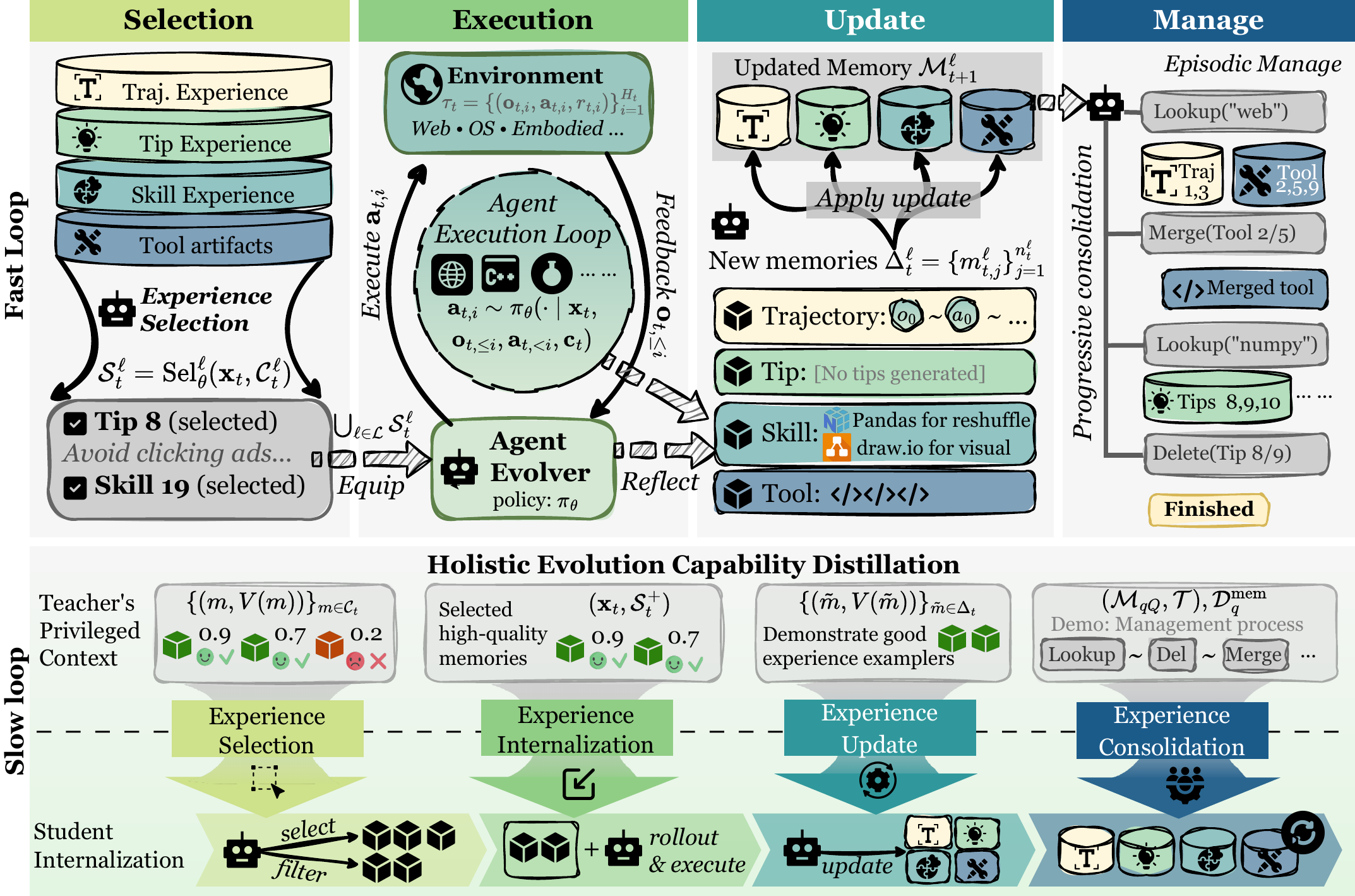}
    \vspace{-1.6em}
    \caption{
    (\textbf{\textit{Top}}) The \textbf{fast loop} lets the agent interact with environments and a four-level memory hierarchy;  (\textbf{\textit{Down}}) the \textbf{slow loop} converts outcome-calibrated hindsight into on-policy self-distillation signals for \ourmethod.
    }
    \label{fig:framework}
    \vspace{-0.8em}
\end{figure*}

\vspace{-0.3em}
\section{Related Work}
\vspace{-0.3em}
\paragraph{Self-Evolving Agents.}
Recent self-evolving agents can be organized around the experience lifecycle defined in this work.
For \ding{182} \emph{experience selection}, prior systems use embedding retrieval~\citep{tan2025prospectretrospectreflectivememory}, utility scoring, learned routers, or policy-based ranking~\citep{li2026memrerankerreasoningawarererankingagent,xu2025singlemultigranularitylongtermmemory} to surface useful memories from growing repositories.
For \ding{183} \emph{experience-grounded execution}, memory- and skill-augmented agents condition policies on retrieved memories~\citep{ferraz2026retrievalaugmentedllmagentslearning,wang2026memexrlscalinglonghorizonllm}, while SFT/RL-based methods internalize experience into policy parameters~\citep{zhang2025agentlearningearlyexperience,liu2026agenticcriticaltraining}.
For \ding{184} \emph{experience writing}, existing methods distill trajectories into reflections~\citep{wu2025evolverselfevolvingllmagents,memp}, reasoning memories, procedural tips~\citep{yue2026memtdensifyingrewardslonghorizon}, executable tools~\citep{zhao2026pyvisionrlforgingopenagentic}, or reusable skills~\citep{skillrl,zhang2026memrlselfevolvingagentsruntime}.
For \ding{185} \emph{experience management}, recent systems study scoring, consolidation~\citep{yue2026memtdensifyingrewardslonghorizon}, forgetting, and architecture-level adaptation~\citep{zhang2025memevolvemetaevolutionagentmemory}.
These works demonstrate the value of experience reuse, but most optimize one or two stages of the lifecycle, whereas our goal is to train a truly holistic agent evolver.

\vspace{-0.3em}
\paragraph{On-Policy Distillation.}
On-policy distillation (OPD) trains a student on its own visited states by querying a teacher for dense supervision, reducing the train-inference mismatch of off-policy distillation~\citep{hinton2015distillingknowledgeneuralnetwork,agarwal2024onpolicydistillationlanguagemodels,li2026rethinkingonpolicydistillationlarge,song2026surveyonpolicydistillationlarge}.
Recent studies have widely explored OPD for improving agent execution, including mathematical reasoning~\citep{wu2026lightningopdefficientposttraining}, knowledge QA~\citep{ye2026onpolicycontextdistillationlanguage}, and tool-use tasks~\citep{zhong2026sodstepwiseonpolicydistillation}.
In contrast, our work uses OPD not merely to strengthen execution, but to jointly cultivate the four capabilities required by a holistic agent evolver.

\vspace{-0.3em}
\section{Methodology}
\vspace{-0.3em}
\subsection{Problem Formulation}
\label{sec:problem-formulation}

We consider a lifelong agent that faces a stream of interactive tasks rather than a single isolated query.
At round $t$, an agent parameterized by $\theta$ receives a task $\mathbf{x}_t \sim \mathcal{D}$, interacts with an environment through observations $\mathbf{o}_{t,i}$ and actions $\mathbf{a}_{t,i}$, and obtains environmental feedback $r_{t,i}$.
The resulting episode is denoted by
\begin{equation}
\tau_t=\{(\mathbf{o}_{t,i},\mathbf{a}_{t,i},r_{t,i})\}_{i=1}^{H_t},
\;
R_t=\sum_{i=1}^{H_t} r_{t,i}
\end{equation}
where $H_t$ is the episode horizon and $R_t$ is the task return.
The per-step signal $r_{t,i}$ may be sparse; when only terminal feedback is available, we place it at the final step.
Unlike standard evaluation that discards $\tau_t$, our agent maintains an evolving repository $\mathcal{M}_t$ of reusable experience, so future behavior depends on how history is selected, used, written, and maintained.
We call such an agent an \textbf{\emph{agent evolver}} if it realizes a closed transition from the current task-memory state $(\mathbf{x}_t,\mathcal{M}_t)$ to both task behavior and an updated repository.
Concretely, it first retrieves a noisy candidate set $\mathcal{C}_t$ and selects a compact context $\mathcal{S}_t$:
\begin{equation}
\mathcal{C}_t=\mathrm{Ret}(\mathbf{x}_t,\mathcal{M}_t),
\;
\mathcal{S}_t=\mathrm{Sel}_{\theta}(\mathbf{x}_t,\mathcal{C}_t)
\end{equation}
where $\mathrm{Ret}$ is a non-parametric retriever, $\mathrm{Sel}_{\theta}$ is the agent's selection behavior, and $\mathcal{S}_t\subseteq\mathcal{C}_t$.
The same agent $\pi_{\theta}$ then executes with $\mathcal{S}_t$ and, after observing the outcome, emits an experience update $\Delta_t$:
\begin{equation}
\begin{gathered}
\small
\tau_t\sim\pi_{\theta}(\cdot\mid\mathbf{x}_t,\mathcal{S}_t),
\;
\Delta_t\sim\pi_{\theta}(\cdot\mid\mathbf{x}_t,\tau_t,R_t,\mathcal{S}_t),\\
\mathcal{M}_{t+1}=\mathcal{M}_t\oplus_{\theta}\Delta_t
\end{gathered}
\end{equation}
where $\Delta_t$ denotes the memories and retention decisions produced from the episode, and $\oplus_{\theta}$ denotes the agent-controlled repository update.
Thus, self-evolution depends on the full experience lifecycle, not merely a memory buffer: $\mathcal{C}_t$ must contain relevant evidence, $\mathcal{S}_t$ must be actionable, $\Delta_t$ reusable, and $\mathcal{M}_{t+1}$ useful as it grows.
The streaming objective is to maximize performance along the task stream while preserving the future utility of the repository:
\begin{equation}
\max_{\theta}\ \liminf_{T\to\infty}\frac{1}{T}\sum_{t=1}^{T}
\mathbb{E}\!\left[R_t+\lambda\,U(\mathcal{M}_{t+1})\right],
\end{equation}
where $U(\mathcal{M}_{t+1})$ denotes the downstream usefulness of the updated repository and $\lambda$ controls how much the agent values future evolvability.
The objective of \ourmethod is to train this closed loop so the agent attains high $R_t$ while improving future repositories.
The following sections instantiate this objective with a fast loop ($\triangleright$ \Cref{sec:fast-evolution}) for online experience-conditioned interaction and a slow loop ($\triangleright$ \Cref{sec:slow-evolution}) that distills privileged hindsight into the unified evolver.

\vspace{-0.1em}
\subsection{Fast Evolution Loop}
\label{sec:fast-evolution}
\vspace{-0.3em}
The fast loop runs online for each task and performs test-time evolution without changing $\theta$.
Its role is to turn $\mathcal{M}_t$ into a small, actionable context before execution, and to turn new episodes back into reusable experiences afterward.

\vspace{-0.3em}
\paragraph{Hierarchical experience memory.}
We organize \ourmethod's $\mathcal{M}_t$ into four complementary tiers.
Let $\mathcal{L}=\{\mathrm{traj},\mathrm{tip},\mathrm{skill},\mathrm{tool}\}$ denote the tier set, where each $\mathcal{M}_t^\ell$ stores memories of type $\ell$:
\begin{equation}
\mathcal{M}_t=
\mathcal{M}_t^{\mathrm{traj}}
\cup \mathcal{M}_t^{\mathrm{tip}}
\cup \mathcal{M}_t^{\mathrm{skill}}
\cup \mathcal{M}_t^{\mathrm{tool}}
\end{equation}
where $\mathcal{M}_t^{\mathrm{traj}}$ preserves episodic evidence, $\mathcal{M}_t^{\mathrm{tip}}$ stores local warnings or heuristics, $\mathcal{M}_t^{\mathrm{skill}}$ abstracts reusable procedures, and $\mathcal{M}_t^{\mathrm{tool}}$ stores executable command or code templates.
The hierarchy is useful because agent experience has different reuse granularity: full trajectories are faithful but verbose, while skills and tools are compact but must be distilled from enough evidence.

\vspace{-0.3em}
\paragraph{Retrieval and selection.}
Given the task $\mathbf{x}_t$ and optional environment metadata $\mathbf{e}_t$, we form a query $\mathbf{z}_t=[\mathbf{x}_t;\mathbf{e}_t]$ and retrieve top-$K$ candidates from each tier by embedding similarity:
\begin{equation}
\small
\mathcal{C}_t^{\ell}
=
\operatorname{TopK}_{m\in\mathcal{M}_t^{\ell}}
\operatorname{sim}\!\left(\phi(\mathbf{z}_t),\phi(m)\right),
\;
\mathcal{C}_t=\bigcup_{\ell\in\mathcal{L}}\mathcal{C}_t^{\ell}
\end{equation}
where $\phi(\cdot)$ is the embedding encoder, $\operatorname{sim}(\cdot,\cdot)$ is cosine similarity, and $\operatorname{TopK}$ returns the $K$ memories with largest scores.
Retrieval alone is intentionally high-recall: it surfaces potentially useful memories, but also brings stale, redundant, or task-mismatched items.
The selector then compresses $\mathcal{C}_t$ into the memory context shown to the agent:
\begin{equation}
\small
\mathcal{S}_t^{\ell}=\mathrm{Sel}_{\theta}^{\ell}(\mathbf{x}_t,\mathcal{C}_t^{\ell}),
\;
\mathcal{S}_t=\bigcup_{\ell\in\mathcal{L}}\mathcal{S}_t^{\ell},
\;
\mathbf{c}_t=\mathrm{Fmt}(\mathcal{S}_t)
\end{equation}
where $\mathrm{Sel}_{\theta}^{\ell}$ selects memories $\mathcal{S}_t^\ell$ from the tier-$\ell$ memory pool $\mathcal{C}_t^\ell$, and $\mathrm{Fmt}$ renders selected memory items into the textual prompt context.
The compact context $\mathbf{c}_t$ is the operational interface between external memory and the acting policy.

\vspace{-0.3em}
\paragraph{Experience-grounded execution and writing.}
During rollout, the actor conditions on $\mathbf{c}_t$ together with the task and interaction history:
\begin{equation}
\mathbf{a}_{t,i}\sim
\pi_{\theta}(\cdot\mid
\mathbf{x}_t,\mathbf{o}_{t,\le i},\mathbf{a}_{t,<i},\mathbf{c}_t)
\end{equation}
where $\mathbf{o}_{t,\le i}$ and $\mathbf{a}_{t,<i}$ denote the partial interaction history before action $\mathbf{a}_{t,i}$.
After the episode ends, the same interaction becomes evidence for future tasks.
The same agent decides which tiers should be updated and how many memories to write to each tier.
Given $\mathbf{x}_t$, $\tau_t$, $R_t$, and $\mathcal{S}_t$, it produces $\Delta_t=\{\Delta_t^\ell\}_{\ell\in\mathcal{L}}$, where each $\Delta_t^\ell$ may be empty or contain multiple new memories:
\begin{equation}
\Delta_t^{\ell}=\{m_{t,j}^{\ell}\}_{j=1}^{n_t^{\ell}},
\;
\mathcal{M}_{t+1}^{\ell}=\mathcal{M}_t^{\ell}\oplus_{\theta}\Delta_t^{\ell}
\end{equation}
The count $n_t^\ell$ is chosen by the agent, so the fast loop can write to any subset of the four tiers rather than following a fixed memory schema.
In addition to per-task writing, the agent periodically performs repository maintenance.
Let $\mathcal{H}_t=\{(\tau_j,R_j,\Delta_j)\}_{j<t}$ denote the logged interaction history.
Every $Q$ tasks, it enters a multi-turn interaction with $\mathcal{H}_{qQ}$ and the current repository, using $\mathrm{lookup}$ plus $\mathrm{merge}(m_i,m_j)$ and $\mathrm{delete}(m_i)$:
\begin{equation}
\eta_q\sim\pi_{\theta}(\cdot\mid\mathcal{M}_{qQ},\mathcal{H}_{qQ},\mathcal{T}),
\;
\mathcal{M}_{qQ}^{+}=\mathcal{M}_{qQ}\oplus_{\theta}\eta_q
\end{equation}
where $q$ indexes maintenance rounds, $qQ$ denotes the task index at which the $q$-th maintenance is triggered, $\mathcal{T}=\{\mathrm{lookup},\mathrm{merge},\mathrm{delete}\}$, $\eta_q$ is the maintenance tool-call trajectory, and $\mathcal{M}_{qQ}^{+}$ is the repository after maintenance.
In parallel, the loop logs retrieved candidates, selected memories, outcomes, created memory ids, and maintenance actions.
These logs are not needed for immediate execution, but they provide the hindsight evidence used by the slow loop to estimate memory value and distill the evolver.

\vspace{-0.1em}
\subsection{Slow Evolution Loop}
\label{sec:slow-evolution}
\vspace{-0.2em}
The fast loop lets the agent accumulate experience, but not necessarily use it well.
An arbitrary agent does not know which memories help, how to ground execution in them, what deserves to become memory, or when a repository update helps future tasks.
The slow loop trains these coupled abilities by turning the agent's own interaction stream into supervision.
The only trustworthy external signal available after each task is the environment feedback $R_t$; the central question is how to propagate this scalar outcome back to selection, execution, writing, and maintenance decisions.

\vspace{-0.3em}
\paragraph{Outcome-calibrated attribution.}
For each memory $m$, we estimate its value only on tasks where $m$ was actually retrieved, so the comparison is candidate-controlled rather than confounded by irrelevant tasks.
Let $g(t)$ denote the task group of round $t$, and define $\Omega_g(m)=\{t:g(t)=g,\,m\in\mathcal{C}_t\setminus \mathcal{S}_t\}\}$ and $\Omega_g^+(m)=\{t:g(t)=g,\,m\in\mathcal{S}_t\}$.
The outcome-calibrated attribution of $m$ is
\begin{equation}
\small
\begin{gathered}
\widehat{A}(m)=\!\!\!
\sum_{g}\!
\rho_g(m)
(
\mathbb{E}_{t\in\Omega_g^+(m)}[R_t]
-
\mathbb{E}_{t\in\Omega_g(m)}[R_t]
)
\end{gathered}
\end{equation}
where $\rho_g(m)=|\Omega_g^+(m)|/(|\Omega_g^+(m)|+|\Omega_g(m)|)$ downweights unreliable selections; empty groups are omitted.
We then convert attribution into a memory score
\begin{equation}
\small
\begin{gathered}
V(m)=\alpha_{\ell(m)}\,\gamma(m)\,\widehat{A}(m),
\\
\gamma(m)=1-\frac{1}{\sqrt{1+N^+(m)}},
\end{gathered}
\end{equation}
where $\ell(m)$ is the tier of $m$, $\alpha_{\ell(m)}$ is a tier prior, $N^+(m)=|\Omega_g^+(m)|$, and $\gamma(m)$ is a confidence factor.
This calibration converts delayed environment feedback into dense hindsight labels: memories with high $V(m)$ become evidence for what should have been selected, used, and preserved, while low-value memories expose noise that the evolver should learn to ignore.

\begin{table*}[!ht]
\centering
\small
\setlength{\tabcolsep}{4pt}
\renewcommand{\arraystretch}{0.98}
\caption{
Main results across self-evolving agent benchmarks.
Each cell reports the official task success metric for the corresponding benchmark subset; higher is better.
For AMA-Bench, CI: Causal Inference, SU: State Updating, SA: State Updating.
\colorbox{bestgreen}{Best} and \colorbox{secondgreen}{second-best} mark the top two results within each backbone group.
}
\vspace{-0.5em}
\label{tab:main-results}
\begin{tabular}{
ll@{\hspace{0.6em}}
cc@{\hspace{0.7em}}
cc@{\hspace{0.7em}}
ccc@{\hspace{0.7em}}
ccc
}
\toprule
\multirow{2}{*}{\textbf{LLM}} &
\multirow{2}{*}{\textbf{Method}} &
\multicolumn{2}{c}{\textbf{LifelongAgentBench}} &
\multicolumn{2}{c}{\textbf{MemoryArena}} &
\multicolumn{3}{c}{\textbf{AMA-Bench}} &
\multicolumn{3}{c}{\textbf{InterCode}} \\
\cmidrule(lr){3-4}\cmidrule(lr){5-6}\cmidrule(lr){7-9}\cmidrule(lr){10-12}
& & \textbf{DB} & \textbf{OS}
& \textbf{Math} & \textbf{Physics}
& \textbf{CI} & \textbf{SU} & \textbf{SA}
& \textbf{Bash} & \textbf{CTF} & \textbf{SQL} \\
\midrule

\multicolumn{2}{l}{{\llmname{Qwen3.5-397B-A17B}}} & 86.00 & 63.00 & 3.98 & 4.65 & 57.21 &  58.73 &  51.41 & 51.34 & 56.00 & 62.74  \\

\multicolumn{2}{l}{{\llmname{Step-3.5-Flash}} (196B)}  & 81.00 & 58.00 & 1.98 & 2.33 & 49.50 & 46.99 &  48.88 & 44.20 & 48.00 & 59.87 \\

\midrule

\multirow{9}{*}{\llmname{Qwen3-4B}}
& No Memory     
& 65.00 & 36.50 & 2.40 & 2.33 & 24.83 & 27.67 & 29.73 & 30.36 & 26.00 & 38.85 \\
& ExpeL         
& 62.50 & 34.50 & 1.98 & 1.74 & 26.17 & 24.88 & 28.49 & 30.80 & \cellcolor{secondgreen}33.00 & 34.08 \\
& AWM           
& 63.00 & 36.00 & 1.69 & \cellcolor{secondgreen}3.49 & 28.36 & 30.14 & \cellcolor{secondgreen}33.49 & 27.68 & \cellcolor{secondgreen}33.00 & 38.54 \\
& Cheatsheet    
& 69.00 & 38.50 & 2.97 & 1.74 & 25.34 & 30.91 & 30.37 & 33.04 & 25.00 & 37.90 \\
& Memp          
& \cellcolor{secondgreen}73.00 & 41.00 & \cellcolor{secondgreen}4.66 & 0.00 & \cellcolor{secondgreen}32.38 & 28.44 & 31.81 & 31.70 & 27.00 & 43.31 \\
& ReasoningBank 
& 70.00 & 38.00 & 3.81 & 1.16 & 27.35 & 29.68 & 30.97 & 31.25 & 25.00 & 43.95 \\
& EvolveR       
& 71.50 & \cellcolor{secondgreen}46.50 & 1.27 & 2.91 & 26.51 & 24.42 & 30.01 & \cellcolor{secondgreen}33.93 & 31.00 & \cellcolor{secondgreen}44.90 \\
& MemEvolve     
& 72.00 & 44.00 & 1.84 & 2.33 & 29.53 & \cellcolor{secondgreen}31.53 & 32.73 & \cellcolor{secondgreen}33.93 & 29.00 & 44.59 \\
& \ourmethod-4B 
& \cellcolor{bestgreen}74.00 & \cellcolor{bestgreen}49.50 & \cellcolor{bestgreen}5.51 & \cellcolor{bestgreen}4.07 & \cellcolor{bestgreen}32.89 & \cellcolor{bestgreen}34.93 & \cellcolor{bestgreen}34.90 & \cellcolor{bestgreen}36.16 & \cellcolor{bestgreen}34.00 & \cellcolor{bestgreen}45.86 \\

\midrule

\multirow{9}{*}{\llmname{Qwen3.5-9B}}
& No Memory     
& 75.50 & 52.50 & 5.51 & 5.23 & 40.10 & 47.14 & 45.63 & 41.52 & 44.00 & 55.73 \\
& ExpeL         
& 74.00 & 46.50 & \cellcolor{secondgreen}8.19 & 8.72 & 41.11 & 43.74 & 44.67 & 40.18 & 50.00 & 52.23 \\
& AWM           
& 72.50  & 44.00 & 7.34 & \cellcolor{secondgreen}11.05 & 43.96 & 48.38 & \cellcolor{secondgreen}48.00 & 39.73 & 49.00 & 51.59 \\
& Cheatsheet    
& 79.50 & 52.50 & 6.64 & 5.23 & 40.77 & 49.30 & 46.19 & 41.96 & 47.00 & 56.05 \\
& Memp          
& 82.00 & 56.00 & 4.80 & 6.40 & \cellcolor{secondgreen}45.81 & 47.14 & 46.59 & 44.20 & 50.00 & 58.28 \\
& ReasoningBank 
& 80.50 & 55.00 & 4.94 & 6.98 & 42.28 & 48.38 & 47.04 & 43.75 & 45.00 & 57.96 \\
& EvolveR       
& \cellcolor{secondgreen}82.50 & 59.50 & 6.07 & 7.56 & 41.78 & 42.81 & 46.11 & 44.64 & 52.00 & 60.51 \\
& MemEvolve     
& 81.00 & \cellcolor{secondgreen}61.00 & 4.24 & 9.88 & 44.30 & \cellcolor{secondgreen}49.77 & 47.20 & \cellcolor{secondgreen}45.98 & \cellcolor{secondgreen}53.00 & \cellcolor{secondgreen}61.15 \\
& \ourmethod-9B 
& \cellcolor{bestgreen}84.50 & \cellcolor{bestgreen}65.00 & \cellcolor{bestgreen}10.88 & \cellcolor{bestgreen}11.63 & \cellcolor{bestgreen}47.32 & \cellcolor{bestgreen}53.94 & \cellcolor{bestgreen}52.92 & \cellcolor{bestgreen}49.55 & \cellcolor{bestgreen}57.00 & \cellcolor{bestgreen}64.01 \\

\bottomrule
\end{tabular}
\vspace{-0.6em}
\end{table*}

\vspace{-0.3em}
\paragraph{Unified hindsight self-distillation.}
We use the calibrated stream to train the same agentic evolver under a single on-policy distillation principle.
Let $\mathcal{K}=\{\mathrm{sel},\mathrm{act},\mathrm{write},\mathrm{maint}\}$.
For each lifecycle decision $k$, the student sees only the public input $z^k$, while the teacher additionally sees a privileged hindsight view $h^k$:
\begin{equation}
\small
\begin{aligned}
(z_t^{\mathrm{sel}},h_t^{\mathrm{sel}})
&=\big((\mathbf{x}_t,\mathcal{C}_t),
\{(m,V(m))\}_{m\in\mathcal{C}_t}\big),\\
(z_t^{\mathrm{act}},h_t^{\mathrm{act}})
&=\big(\mathbf{x}_t,(\mathcal{S}_t^{+},\tau_t^{+})\big),\\
(z_t^{\mathrm{write}},h_t^{\mathrm{write}})
&=\big((\mathbf{x}_t,\tau_t,R_t,\mathcal{S}_t),
\{(\tilde{m},V(\tilde{m}))\}_{\tilde{m}\in\Delta_t}\big),\\
(z_q^{\mathrm{maint}},h_q^{\mathrm{maint}})
&=\big((\mathcal{M}_{qQ},\mathcal{H}_{qQ},\mathcal{T}),
\mathcal{D}_{q}^{\mathrm{mem}}\big)
\end{aligned}
\end{equation}
where $\mathcal{S}_t^{+}=\{m\in\mathcal{S}_t:V(m)>0\}$ denotes valuable selected memories, $\tau_t^{+}$ is a successful trajectory from the same task group, and $\mathcal{D}_{q}^{\mathrm{mem}}=\{(m,V(m),\gamma(m),\nu(m))\}_{m\in\mathcal{M}_{qQ}}\cup\{\kappa(m_i,m_j)\}_{i,j}$ contains memory value, confidence, usage statistic $\nu(m)$, and redundancy score $\kappa(m_i,m_j)$.
Thus, \ding{182} selection sees the value of every retrieved candidate; \ding{183} execution internalizes useful selected memories by asking the student to solve without memory; \ding{184} writing sees which produced memories later become valuable; and \ding{185} maintenance sees calibrated diagnostics for producing merge/delete tool-call trajectories.
For each $(z^k,h^k)$, the student first samples an on-policy output $\hat{\mathbf{y}}^k$ under the deployment condition, and the teacher is evaluated on the same student prefixes:
\begin{equation}
\begin{gathered}
\hat{\mathbf{y}}^k\sim p_{\theta}^{S}(\cdot\mid z^k),\;
p_{\theta,n}^{S}=p_{\theta}^{S}(\cdot\mid z^k,\hat{\mathbf{y}}_{<n}^k),
\\
p_{\bar{\theta},n}^{T}=p_{\bar{\theta}}^{T}(\cdot\mid z^k,h^k,\hat{\mathbf{y}}_{<n}^k)
\end{gathered}
\end{equation}
where $p_{\theta}^{S}$ is the student policy, $p_{\bar{\theta}}^{T}$ is the same evolver under privileged conditioning with stop-gradient parameters $\bar{\theta}$, and $\hat{\mathbf{y}}_{<n}^k$ is the student-generated prefix.
Let $\delta_{k,n}$ denote the corresponding token-level discrepancy:
\begin{equation}
\delta_{k,n}
=
D_{\mathrm{tok}}\!\left(
\operatorname{sg}[p_{\bar{\theta},n}^{T}]
\;\Vert\;
p_{\theta,n}^{S}
\right)
\end{equation}
where $D_{\mathrm{tok}}$ is full-vocabulary token-level KL divergence, and $\operatorname{sg}[\cdot]$ blocks gradients through the teacher distribution.
The unified slow-loop objective is the token-level distillation loss over student-visited prefixes:
\begin{equation}
\small
\begin{aligned}
\mathcal{L}_{\mathrm{slow}}(\theta)=
\sum_{k\in\mathcal{K}}\;
\mathbb{E}_{\substack{(z^k,h^k)\sim d_{\pi_{\theta}}^k\\
\hat{\mathbf{y}}^k\sim p_{\theta}^{S}(\cdot\mid z^k)}}
\bigg[
\frac{1}{L_k}\sum_{n=1}^{L_k}\delta_{k,n}
\bigg]
\end{aligned}
\end{equation}
where $d_{\pi_{\theta}}^k$ is the on-policy distribution of decision inputs for lifecycle decision $k$.
Selection, execution, writing, and maintenance are therefore supervised as four views of one experience lifecycle: the teacher explains what the agent should have selected, how it should have acted with useful experience, and what it should have written or retained after observing the outcome.
After distillation, only the student-facing behavior is deployed back into the fast loop, so the agent can exercise these evolver abilities without privileged feedback at test time.

\begin{algorithm}[!t]\small
\DontPrintSemicolon
\SetAlgoLined
\SetAlgoNoEnd
\LinesNumbered
\KwIn{Task stream $\{\mathbf{x}_t\}$; memory repository $\mathcal{M}_0$; evolver policy $\pi_\theta$; maintenance period $Q$}
\KwOut{Deployable evolver $\pi_\theta^S$ and evolved repository $\mathcal{M}$}

\tcc{\algtt{Fast loop: online experience use}}
\For{\rm{each task } $\mathbf{x}_t$}{
    Form query $\mathbf{z}_t$ from task and environment\;
    \For{$\ell \in \mathcal{L}$}{
        Retrieve $\mathcal{C}_t^\ell$ from $\mathcal{M}_t^\ell$ and select $\mathcal{S}_t^\ell\subseteq\mathcal{C}_t^\ell$\;
    }
    $\mathcal{S}_t \leftarrow \bigcup_{\ell\in\mathcal{L}}\mathcal{S}_t^\ell$, \quad $\mathbf{c}_t \leftarrow \mathrm{Fmt}(\mathcal{S}_t)$\;
    Roll out $\pi_\theta$ with $\mathbf{c}_t$ to obtain $(\tau_t,R_t)$\;
    \For{$\ell \in \mathcal{L}$}{
        Write $\Delta_t^\ell$ and update $\mathcal{M}_{t+1}^\ell \leftarrow \mathcal{M}_{t}^\ell \oplus_\theta \Delta_t^\ell$\;
    }
    \If{$t \bmod Q = 0$}{
        $q \leftarrow t/Q$\;
        Produce maintenance trajectory $\eta_q$ with $\{\mathrm{lookup},\mathrm{merge},\mathrm{delete}\}$ and update $\mathcal{M}_{qQ}^{+}$\;
    }
    Log $(\mathcal{C}_t,\mathcal{S}_t,R_t,\Delta_t)$ and any $\eta_q$\;
}

\tcc{\algtt{Slow loop: unified distill}}
\For{\rm{each logged memory } $m$}{
    Estimate $V(m)$ from logged outcomes\;
}
\For{$k \in \{\mathrm{sel},\mathrm{act},\mathrm{write},\mathrm{maint}\}$}{
    Construct $(z^k,h^k)$, sample $\hat{\mathbf{y}}^k\sim p_\theta^S(\cdot\mid z^k)$, and minimize $\mathcal{L}_{\mathrm{slow}}$\tcp{\algtt{OPD}}
}
\Return Qualified agent evolver $\pi_\theta^S$
\caption{\small Workflow of \ourmethod.}
\label{algo:opd_evolver}
\end{algorithm}

\vspace{-0.3em}
\section{Experiments}
\vspace{-0.3em}
\subsection{Experiment Setup}
\label{sec:experiment-setup}
\vspace{-0.3em}

\begin{table*}[!t]
\centering
\begin{minipage}[t]{0.56\linewidth}
\centering
\caption{
Comparison with training-based agent improvement methods.
Each cell reports the official task success metric.
}
\vspace{-0.8em}
\label{tab:training-comparison}
\footnotesize
\setlength{\tabcolsep}{2.5pt}
\renewcommand{\arraystretch}{1}
\resizebox{\linewidth}{!}{
\begin{tabular}{lccccccc}
\toprule
\multirow{2}{*}{\textbf{Method}} &
\multicolumn{3}{c}{\textbf{MiniHack}} &
\multicolumn{3}{c}{\textbf{InterCode}} \\
\cmidrule(lr){2-4}\cmidrule(lr){5-7}
& \textbf{Room} & \textbf{Maze} & \textbf{KeyRoom} 
& \textbf{Bash} & \textbf{CTF} & \textbf{SQL} \\
\midrule
Vanilla          & 80.39 & 19.61 & 0.00 & 55.73 & 44.00 & 41.25 \\
SFT              & 82.35 & 17.65 & 0.00 & 59.87 & 49.00 & 44.64 \\
GRPO             & \cellcolor{bestgreen}100.00 & 23.53 & 3.92 & \cellcolor{secondgreen}63.69 & \cellcolor{secondgreen}58.00 & 47.77 \\
Skill0           & 94.12 & \cellcolor{secondgreen}25.49 & 3.92 & 62.10 & 54.00 & 47.32  \\
MemRL         & 96.08 & 19.61 & 1.96 & 61.15 & 50.00 & 44.20  \\
Complementary RL & 96.88 & 20.85 & \cellcolor{secondgreen}5.16 & 63.20 & 55.00 & \cellcolor{secondgreen}48.10  \\
\ourmethod       & \cellcolor{secondgreen}98.04 & \cellcolor{bestgreen}27.45 & \cellcolor{bestgreen}9.80 & \cellcolor{bestgreen}64.01 & \cellcolor{bestgreen}59.00 & \cellcolor{bestgreen}49.55  \\
\bottomrule
\end{tabular}
}
\end{minipage}
\hfill
\begin{minipage}[t]{0.41\linewidth}
\centering
\caption{
Ablation study on InterCode (Bash/CTF/SQL) with \ourmethod-4B.
``Writing Distill.'' denotes the exclusion of self-distilling experience writing capability.
}
\vspace{-0.5em}
\label{tab:ablation-intercode}
\footnotesize
\setlength{\tabcolsep}{3.5pt}
\renewcommand{\arraystretch}{1}
\resizebox{\linewidth}{!}{
\begin{tabular}{lccc}
\toprule
\textbf{Variant} & \textbf{Bash} & \textbf{CTF} & \textbf{SQL} \\
\midrule
\ourmethod-4B & \cellcolor{bestgreen}36.16 & \cellcolor{bestgreen}34.00 & \cellcolor{bestgreen}45.86 \\
\quad w/o Slow Evolution & 32.14 & 28.00 & 39.17 \\
\quad w/o Memory Attribution & 31.20 & 26.69 & 38.50 \\
\quad w/o Selection & 35.27 & \cellcolor{secondgreen}32.00 & 42.04 \\
\quad w/o Writing Distill. & 34.38 & 29.00 & 41.08 \\
\quad w/o Maintenance & \cellcolor{secondgreen}35.30 & 30.10 & \cellcolor{secondgreen}43.51 \\
\bottomrule
\end{tabular}
}
\end{minipage}
\vspace{-0.5em}
\end{table*}

\paragraph{Training setup.}
We train \ourmethod on heterogeneous agentic experience from Agent World Model~\citep{wang2026agentworldmodelinfinity}, nvidia/Nemotron-Terminal-Corpus~\citep{pi2026dataengineeringscalingllm}, and EnvScaler~\citep{song2026envscalerscalingtoolinteractiveenvironments}, which are disjoint from all evaluation benchmarks (see \Cref{app:train}).
Unless specified otherwise, we use \llmname{Qwen3-4B-Instruct-2507}~\citep{yang2025qwen3technicalreport} and \llmname{Qwen3.5-9B}~\citep{qwenQwenStudio} as backbones, \llmname{Qwen3-Embedding-0.6B} for retrieval, retrieve $50$ candidates, set $Q=30$, keep at most $20$ memories in privileged teacher contexts, filter supervision with minimum score $0.01$. More parameter setups are detailed in \Cref{app:train}.

\vspace{-0.4em}
\paragraph{Baselines.}
We compare with \ding{110} \textbf{memory-augmented agents}: ExpeL~\citep{expel}, AWM~\citep{wang2024agentworkflowmemory}, Cheatsheet~\citep{suzgun2025dynamiccheatsheettesttimelearning}, MemP~\citep{fang2025mempexploringagentprocedural}, ReasoningBank~\citep{ouyang2025reasoningbankscalingagentselfevolving}, EvolveR~\citep{wu2025evolverselfevolvingllmagents}, and MemEvolve~\citep{zhang2025memevolvemetaevolutionagentmemory}; and \ding{110} \textbf{training-based methods}: SFT, GRPO~\citep{deepseekmath}, Skill0~\cite{skill0}, MemRL~\citep{memrl} and Complementary RL~\citep{comp-rl}.
For fairness, all memory-based methods, including \ourmethod, start evaluation with an empty experience repository and accumulate memory only from the evaluation stream (see details in \Cref{app:evaluation-protocol}).

\vspace{-0.4em}
\paragraph{Evaluation benchmarks.}
We evaluate on five benchmarks: \textbf{LifelongAgentBench}~\citep{zheng2025lifelongagentbenchevaluatingllmagents} with \textit{database (DB)} and \textit{operating system (OS)}; \textbf{MemoryArena}~\citep{he2026memoryarenabenchmarkingagentmemory} with \textit{Math} and \textit{Physics} subsets; \textbf{AMA-Bench}~\citep{zhao2026amabenchevaluatinglonghorizonmemory} with \textit{Causal Inference}, \textit{State Updating}, and \textit{State Abstraction}; \textbf{InterCode}~\citep{yang2023intercode} with \textit{Bash}, \textit{CTF}, and \textit{SQL} subsets; and embodied environment \textbf{MiniHack}~\citep{samvelyan2021minihackplanetsandboxopenended}. Benchmark details are in \Cref{app:benchmark}.

\vspace{-0.2em}
\subsection{Main Results}
\vspace{-0.2em}
\paragraph{Comparison with Memory Systems}
We first compare \ourmethod with seven mainstream self-evolving memory systems as well as several larger models in \Cref{tab:main-results}.
\ourmethod achieves the best result on all $10$ subsets among same-backbone memory baselines for both 4B and 9B, with clear gains over the strongest memory baseline on OS ($65.00\%$ vs.\ $61.00\%$), AMA-SA ($52.92\%$ vs.\ $48.00\%$), and InterCode-CTF ($57.00\%$ vs.\ $53.00\%$) for 9B.
Notably, \ourmethod-9B challenges its much larger counterparts: \ourmethod-9B exceeds \llmname{Step-3.5-Flash} (196B) on $9/10$ subsets and surpasses \llmname{Qwen3.5-397B-A17B} on $6/10$ subsets, including AMA-SA ($52.92\%$ vs.\ $51.41\%$), InterCode-CTF ($57.00\%$ vs.\ $56.00\%$), and SQL ($64.01\%$ vs.\ $62.74\%$), showing that lifecycle-level evolution can make a compact agent competitive with far larger counterparts.

\vspace{-0.2em}
\paragraph{Comparison with Training-based Methods}
\Cref{tab:training-comparison} shows that \ourmethod achieves the best average performance among training-based methods, winning on $5/6$ subsets.
Compared with GRPO, \ourmethod improves Maze ($27.45\%$ vs.\ $23.53\%$), KeyRoom ($9.80\%$ vs.\ $3.92\%$), Bash ($64.01\%$ vs.\ $63.69\%$), CTF ($59.00\%$ vs.\ $58.00\%$), and SQL ($49.55\%$ vs.\ $47.77\%$).
Against Complementary RL, \ourmethod further improves the hardest MiniHack subsets, including Maze ($27.45\%$ vs.\ $20.85\%$) and KeyRoom ($9.80\%$ vs.\ $5.16\%$).
It also consistently outperforms MemRL on both MiniHack and InterCode, with a $5.35$-point gain on SQL.
These results suggest that \ourmethod learns beyond task-level reward fitting, yielding a more transferable mechanism for selecting, internalizing, and reusing experience.

\vspace{-0.1em}
\subsection{Ablation Study}
\vspace{-0.2em}
We ablate five components of \ourmethod in \Cref{tab:ablation-intercode}: \ding{182} w/o Slow Evolution removes on-policy distillation, \ding{183} w/o Memory Attribution replaces outcome-calibrated attribution with a simpler memory-frequency calibrated score, \ding{184} w/o Selection directly chooses the top-5 memories by similarity, \ding{185} w/o Writing Distillation removes supervision for reusable memory writing, and \ding{186} w/o Maintenance disables repository update training.
All variants degrade InterCode performance, with the largest drop from removing memory attribution (average $38.67\%\!\to\!32.13\%$; Bash/CTF/SQL decrease by $4.96\%$, $7.31\%$, and $7.36\%$).
Removing slow evolution is similarly harmful (average $33.10\%$), showing that privileged hindsight must be distilled into the deployable policy.
Selection, writing, and maintenance also matter: similarity-only selection reduces SQL from $45.86\%$ to $42.04\%$, and removing writing distillation lowers CTF from $34.00\%$ to $29.00\%$.
Together, these results show that calibrated attribution, learned selection, memory writing, and repository maintenance must be trained jointly for reliable self-evolution.

\begin{figure}[!t]
\centering
\includegraphics[width=\linewidth]{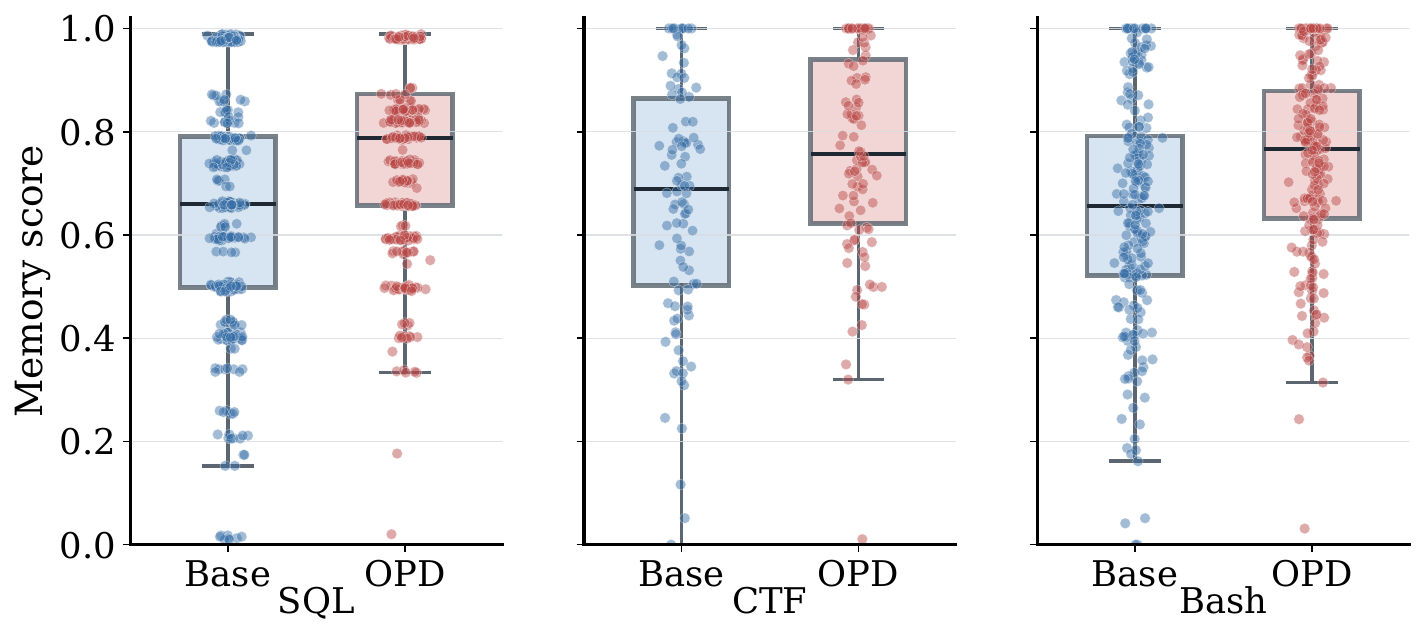}
\vspace{-1.5em}
\caption{
Distribution of calibrated memory scores for memories selected by the original \llmname{Qwen3.5-9B} and \ourmethod-9B after selection distillation.
}
\label{fig:selection-distill}
\vspace{-0.2em}
\end{figure}

\subsection{Framework Analysis}

\paragraph{Impact of Selection Distillation.}
We compare the memory scores of items selected by the original 9B backbone and the trained \ourmethod-9B selector.
As shown in \Cref{fig:selection-distill}, selection distillation consistently raises the median selected-memory score, from $0.66/0.69/0.66$ to $0.79/0.76/0.76$ on SQL/CTF/Bash.
The lower quartile also moves upward from $0.50$ to above $0.62$ across the three subsets, suggesting that \ourmethod reduces low-utility retrieval noise rather than merely shifting a few high-scoring outliers.

\begin{figure}[!t]
\centering
\includegraphics[width=\linewidth]{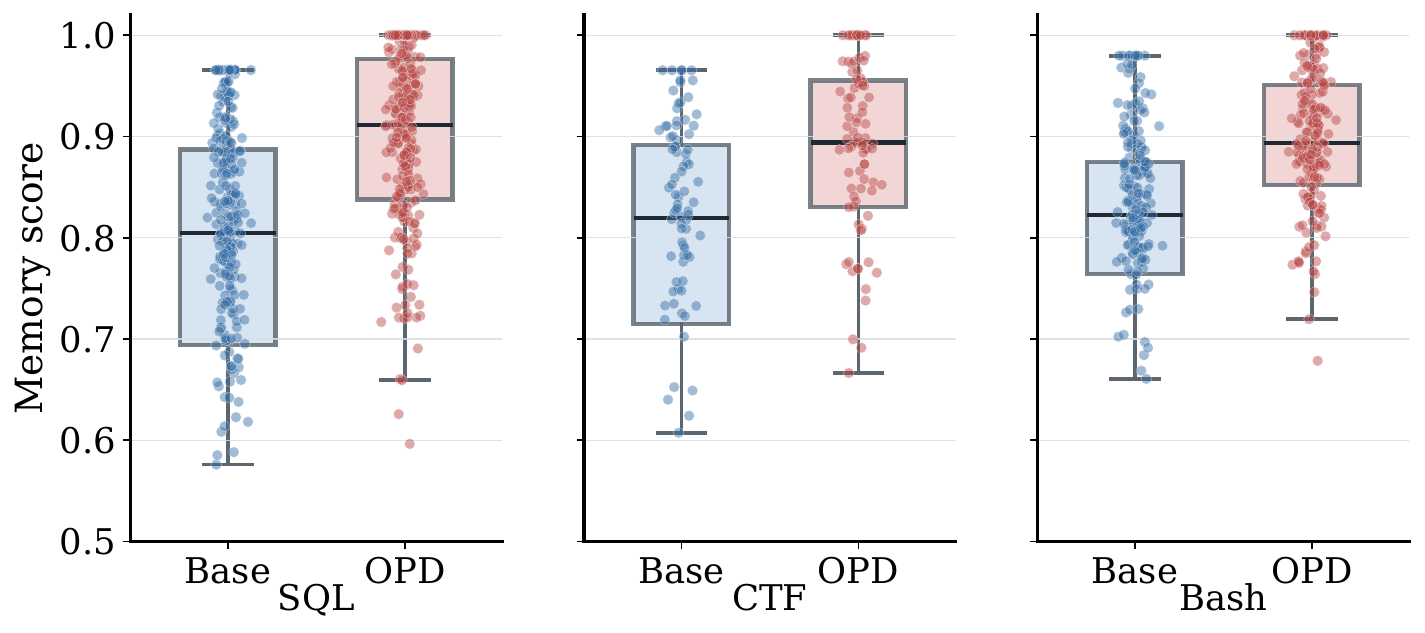}
\vspace{-1.8em}
\caption{
Distribution of calibrated memory scores for memories written by the original \llmname{Qwen3.5-9B} and \ourmethod-9B after writing distillation.
}
\label{fig:writing-distill}
\vspace{-0.4em}
\end{figure}

\paragraph{Impact of Writing Distillation}
We next examine whether the trained evolver writes memories with higher future utility.
\Cref{fig:writing-distill} compares the calibrated score distribution of memories produced by vanilla \llmname{Qwen3.5-9B} and by \ourmethod-9B.
Writing distillation increases the median score from $0.80/0.82/0.82$ to $0.91/0.90/0.89$ on SQL/CTF/Bash, and lifts the lower quartile to $0.83$ or higher.
This tighter high-score distribution indicates that \ourmethod writes memories that are not only better formatted, but more reliably useful for downstream tasks.

\begin{figure}[!t]
\centering
\includegraphics[width=\linewidth]{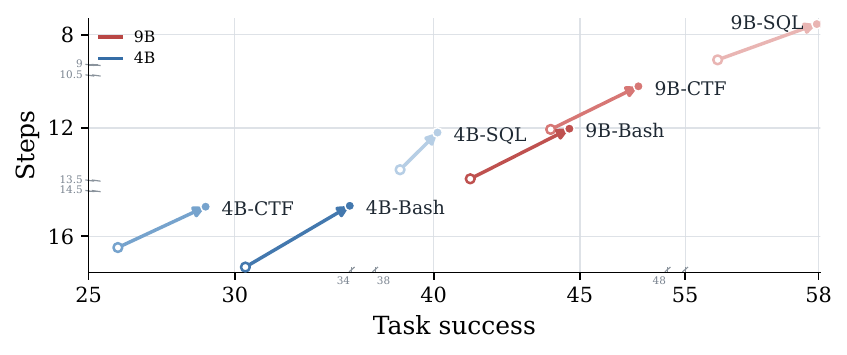}
\vspace{-1.8em}
\caption{
Experience internalization effect.
Each arrow starts from the vanilla backbone and ends at the corresponding \ourmethod policy; rightward movement indicates higher task success and upward movement in the inverted axis indicates fewer steps.
}
\label{fig:experience-internalization}
\vspace{-0.5em}
\end{figure}

\vspace{-0.2em}
\paragraph{Impact of Experience Internalization}
Finally, we test whether slow-loop distillation internalizes useful experience into the policy.
\Cref{fig:experience-internalization} compares each vanilla backbone with its trained \ourmethod counterpart (w/o external $\mathcal{M}$; merely for task execution) in terms of task success and execution steps.
Every arrow moves toward \textit{higher success} and \textit{fewer steps}: \ourmethod raises success by about $3$--$4$ points for 4B and $3$--$7$ points for 9B on Bash/CTF/SQL, while reducing trajectories by up to $2.5$ steps.
This suggests that \ourmethod converts high-value memories into more direct and efficient behavior, rather than merely retrieving better context at inference time.

\subsection{Case Study}
\begin{figure}[!t]
\centering
\includegraphics[width=\linewidth]{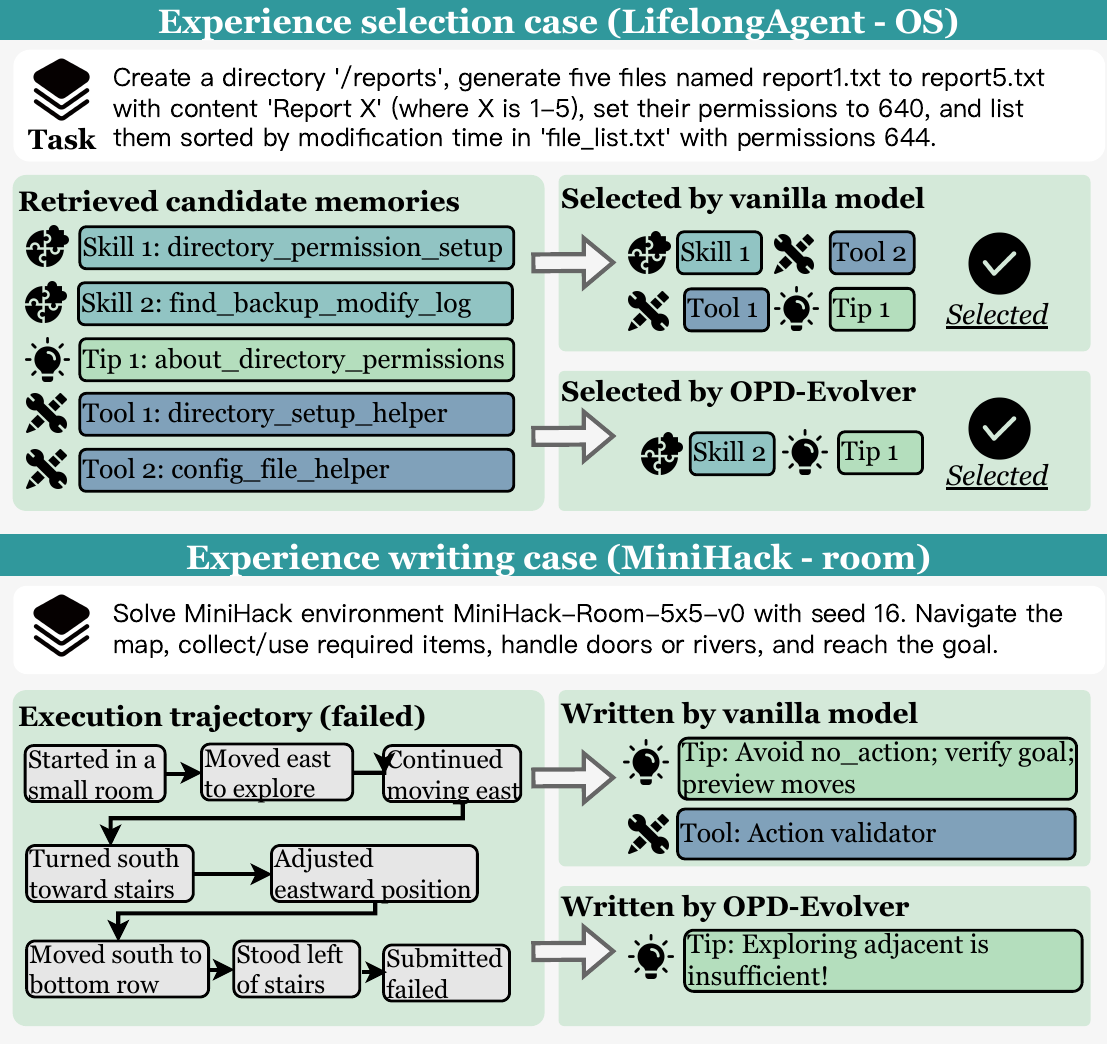}
\vspace{-1.5em}
\caption{
Case study of experience selection and writing in \ourmethod.
(\textbf{\textit{Top}}) on LifelongAgentBench-OS, \ourmethod selects a compact task-matched subset from retrieved memories.
(\textbf{\textit{Bottom}}) on MiniHack-Room, \ourmethod writes a more causal and reusable memory from a failed trajectory.
}
\label{fig:case}
\vspace{-0.4em}
\end{figure}

\Cref{fig:case} illustrates how slow-loop training makes both memory selection and memory writing more precise.
For the LifelongAgentBench-OS task, the vanilla model selects broad directory and configuration memories, while \ourmethod keeps only the modification-log skill and permission tip that directly support creating files, setting modes, and producing the sorted file list.
For the MiniHack-Room failure, the vanilla model writes generic advice such as verifying the goal and adding an action validator, but the trajectory shows a sharper error: the agent stopped after local east/south exploration without sufficiently expanding adjacent states around the stairs.
\ourmethod therefore writes the reusable causal tip ``Exploring adjacent is insufficient,'' showing that the trained evolver learns not only to retrieve relevant experience, but also to turn failures into compact future-facing memory. More case studies are at \Cref{app:case}.

\vspace{-0.1em}
\section{Conclusion}
\vspace{-0.4em}
In this work, we present \ourmethod, a slow-fast co-evolution framework that combines online memory-conditioned interaction with outcome-calibrated attribution and privileged on-policy self-distillation.
Across diverse evaluation benchmarks, \ourmethod consistently improves compact open models over memory-augmented and training-based baselines, showing that evolver abilities learned from heterogeneous streams can transfer beyond the training environments.
Together, these results suggest a shift from building agents that merely store experience toward cultivating agents that can continuously transform experience into their own capacity for evolution.

\bibliography{custom}

\appendix

\section{Related Work Discussion}\label{app:related}

We further discuss several closely related contemporary works, many of which study skill-based experience internalization.
SKILL0~\citep{skill0} and OpenClaw-RL~\citep{wang2026openclawrltrainagentsimply} use RL or OPD to internalize agent execution behaviors, while our focus additionally includes how experience is selected, written, and maintained.
SkillRL~\citep{skillrl}, Trace2Skill~\citep{ni2026trace2skilldistilltrajectorylocallessons}, and Skill-R1~\citep{vishe2026skillr1agentskillevolution} mainly improve skill or lesson generation, corresponding most closely to the experience-writing part of our lifecycle.
Skill-SD~\citep{wang2026skillsdskillconditionedselfdistillationmultiturn} uses skills as privileged context during distillation, emphasizing experience-guided execution.
SLIM~\citep{shen2026dynamicskilllifecyclemanagement} and Skill1~\citep{shi2026skill1unifiedevolutionskillaugmented} take broader views of skill lifecycle management.
In comparison, \ourmethod studies a unified evolver policy over multiple experience forms, including trajectories, tips, skills, and tools, and evaluates its cross-domain generalization across heterogeneous agent benchmarks.

\section{Potential Risks}
We do not observe direct evidence that \ourmethod introduces qualitatively new social risks beyond those already associated with LLM-based agents.
However, because the framework improves an agent's ability to reuse experience and maintain memory, standard concerns around unsafe tool use, privacy leakage, biased behavior, and unintended task generalization still require careful attention.
In practice, deployments should follow the same safeguards expected for agentic LLM systems, including controlled environments, memory auditing, access restrictions, and human oversight when actions may affect real users or external systems.

\section{Benchmark Details}
\label{app:benchmark}

This section provides additional details on the benchmarks used for
evaluation, including their subsets, test-set sizes, environment interfaces, and action
spaces. 

\paragraph{LifelongAgentBench.}
LifelongAgentBench~\citep{zheng2025lifelongagentbenchevaluatingllmagents} is a
multi-domain lifelong agent benchmark. We evaluate on its database (DB) and
operating system (OS) subsets, with 100 test tasks for each subset. The
DB subset consists of MySQL tasks, where the agent receives a natural-language
instruction together with database context and acts through SQL execution
followed by final submission. The OS subset consists of Ubuntu-container tasks,
where the agent receives a natural-language instruction and interacts with the
environment through bash commands followed by final submission. These two subsets
evaluate whether the agent can reuse memory for long-horizon executable tasks
that require precise environment operations.

\paragraph{MemoryArena.}
MemoryArena~\citep{he2026memoryarenabenchmarkingagentmemory} evaluates memory
over multi-session reasoning tasks. We evaluate on the Math and Physics subsets,
which contain 354 and 86 test tasks, respectively. Each task provides
shared background information and a sequence of question-answer sessions. The
agent must answer the current question while preserving and reusing information
from prior sessions. Unlike interactive control benchmarks, MemoryArena does not
require live environment actions; instead, it tests whether the agent can
maintain, retrieve, and apply useful memory across sessions.

\paragraph{AMA-Bench.}
AMA-Bench~\citep{zhao2026amabenchevaluatinglonghorizonmemory} evaluates
long-horizon memory over completed agent trajectories. We use 208
episodes, containing 2,496 question-answer pairs in total. Following the
benchmark taxonomy, the QA pairs are divided into four categories: Recall
(839 pairs), Causal Inference (596 pairs), State Updating (647 pairs), and State
Abstraction (414 pairs). In our evaluation, we focus on the memory-intensive
Causal Inference, State Updating, and State Abstraction categories. Each example
contains an action-observation trajectory and a question about the trajectory.
Causal Inference requires the agent to reason about why events occurred, State
Updating requires tracking changes across the trajectory, and State Abstraction
requires summarizing or abstracting the relevant state. The agent does not
control a live environment in AMA-Bench; it reads the trajectory and produces an
answer, optionally using retrieved memory.

\paragraph{InterCode.}
InterCode~\citep{yang2023intercode} provides interactive coding benchmarks in
isolated execution environments. We evaluate on the SQL, Bash, and CTF subsets,
which contain 314, 224, and 100 test tasks, respectively. In SQL, the
agent solves text-to-SQL tasks by issuing SQL queries against a database. In
Bash, the agent solves natural-language shell tasks by executing bash commands.
In CTF, the agent solves capture-the-flag challenges by executing commands and
attempting flags. In all three subsets, the agent interacts through a JSON action
interface and terminates with final submission. For Bash and SQL, grading is
based on the output of the last executed command or query.

\paragraph{MiniHack.}
MiniHack~\citep{samvelyan2021minihackplanetsandboxopenended} is a suite of
procedurally generated grid-world navigation environments built on NetHack. We
evaluate on Room, Maze, and KeyRoom, with 51 test tasks for each
environment. Room requires the agent to reach the goal in a small open room;
Maze requires systematic exploration in a larger maze; and KeyRoom requires the
agent to locate a key, open a locked door, and then reach the goal. We render
each environment as a text-based dungeon view using NetHack glyph conventions,
such as \texttt{@} for the agent, \texttt{>} for stairs or the goal, and
\texttt{.} for floor tiles. At each step, the agent receives the ASCII map, the
latest environment message, inventory text, and step/reward statistics. The
agent acts through a unified JSON action interface with compass moves,
object-interaction actions such as \texttt{pickup}, \texttt{apply},
\texttt{open}, and \texttt{search}, navigation actions such as
\texttt{climb\_up} and \texttt{wait}, and final \texttt{submit}.

\subsection{Evaluation Protocol}
\label{app:evaluation-protocol}

For all memory-based methods, evaluation starts from an empty experience
repository $\mathcal{M}_0=\emptyset$.
The agent then solves tasks sequentially in a streaming order, so its repository
can grow only from trajectories, memories, and feedback observed during the
current evaluation stream.
After each task, the environment returns task-level feedback indicating whether
the attempt succeeds or fails, which is used for memory update and attribution.
The agent does not receive the ground-truth answer or an expert solution after
the task; it only observes the same environment feedback available to all
compared methods.
For the training-based comparison in \Cref{tab:training-comparison}, we
additionally train all compared methods on MiniHack using the same training data
and evaluate them on the same test tasks to ensure a fair comparison.

\section{Training Details}\label{app:train}

\subsection{Training Data}
\label{app:training-data}

We construct the training data from 7,000 interactive agent tasks: 3,000 from
Agent World Model (AWM)~\citep{wang2026agentworldmodelinfinity}, 2,000 from
nvidia/Nemotron-Terminal-Corpus~\citep{pi2026dataengineeringscalingllm}, and
2,000 from EnvScaler~\citep{song2026envscalerscalingtoolinteractiveenvironments}.
These sources cover diverse executable and tool-interactive scenarios, exposing
\ourmethod to heterogeneous trajectories for memory selection, memory writing,
memory-conditioned execution, and repository maintenance.

\paragraph{Agent World Model (AWM).}
AWM provides synthetic multi-turn tool-use environments with executable,
code-driven state transitions and database-backed application scenarios.
We use it to collect trajectories with reliable task-level outcomes and rich
tool interactions.

\paragraph{nvidia/Nemotron-Terminal-Corpus.}
Nemotron-Terminal-Corpus provides terminal-style interaction data covering
command execution, shell operations, and multi-step problem solving in textual
environments.
It complements AWM by exposing the agent to command-line workflows and
environment feedback patterns.

\paragraph{EnvScaler.}
EnvScaler provides scalable tool-interactive environments for collecting diverse
agent trajectories.
We use it to broaden the training distribution beyond a single benchmark style
and to expose \ourmethod to varied interaction structures.

Note that, for the training-based comparison in \Cref{tab:training-comparison}, we
additionally train all compared methods on MiniHack using the same training data
and evaluate them on the same test tasks to ensure a fair comparison.

\subsection{Training Parameters}
\label{app:training-parameters}

This section summarizes the training and rollout hyperparameters used in our
experiments. 

\paragraph{Shared Settings}
Unless otherwise specified, we use Qwen3.5-9B and Qwen3-4B backbones with
\texttt{bf16} precision. We set the maximum prompt length to 8192 tokens. 
For GRPO training, we use temperature 1.0 and top-$p$ 1.0. For
\ourmethod distillation data generation, we use temperature 1.0 and top-$p$
0.95. Across interactive environments, the maximum episode length is 40 steps
unless otherwise specified.

\paragraph{SFT}
For the SFT baseline, we fine-tune the base model with LoRA. We use a learning
rate of $1\times10^{-5}$, per-device batch size 2, gradient accumulation 4, and
train for 3 epochs. The LoRA rank is 32, the LoRA scaling factor is 64, and the
LoRA dropout is 0.05.

\paragraph{GRPO}
For the GRPO baseline, we use the VERL framework and initialize the policy
directly from the base model. We use group
size 8 and rollout batch size 2. The learning rate is $1\times10^{-6}$, and we
train for one epoch. We apply actor-side KL regularization with coefficient
0.001, but do not include the KL term in the reward. The reward is based on the
environment cumulative reward. Invalid actions receive a penalty of $-0.1$, and
premature submission receives a penalty of $-0.2$.

\paragraph{MemRL}
For MemRL, we follow its non-parametric runtime memory reinforcement learning
setup, where learning is performed through episodic memory updates rather than
model weight updates. We use Qwen3.5-9B as the LLM backend and
Qwen3-Embedding-0.6B as the embedding model. We retrieve at most 3 memories per
query. The memory construction, retrieval, and update strategies are
\texttt{proceduralization}, \texttt{query}, and \texttt{adjustment},
respectively. For Q-value updates, we use a success reward of $+1.0$ and a
failure reward of $0.0$.  Other MemRL
reinforcement-learning hyperparameters are kept at the package defaults.

\paragraph{SkillZero}
For SkillZero, we use its skill-conditioned GRPO training setup implemented with
the VERL framework. Skill conditioning is enabled during
training. We initialize SkillZero with a MiniHack-specific skill set and use a
curriculum schedule that starts with all skills, switches to core skills after
step 100, and removes skills after step 200. The rollout group size is 4, the
learning rate is $1\times10^{-6}$, and the batch size is 2. We train SkillZero
for one GRPO epoch. Invalid actions receive a penalty of $-0.1$.

\paragraph{ComplementaryRL}
ComplementaryRL is implemented with the ROLL framework under
\texttt{MODE=complementaryrl}. It uses a fully asynchronous actor-memory
pipeline with three non-shared modules: a Qwen3.5-9B actor, a
Qwen3-4B-Thinking-2507 memory model, and a Qwen3-4B-Thinking-2507 memory actor,
with Qwen3-Embedding-0.6B for retrieval. The actor is trained with GRPO, while
the memory actor is trained with CISPO and REINFORCE-style advantage estimation.
We use learning rate $1\times10^{-6}$, rollout batch size 128, group size 1,
gradient accumulation 64, and 512 training steps. Actor rollouts use temperature
0.99, top-$p$ 0.99, and at most 4096 new tokens per step. The memory system
stores trajectory memories, retrieves the top-1 memory by embedding similarity,
uses FIFO updates, and caps the memory pool at 20,000 items. KL regularization
is disabled, and episodes are limited to 80 actions per trajectory.

\paragraph{\ourmethod}
For \ourmethod, the memory pool contains four types of memories: skills, tips,
tools, and trajectories. During teacher-side retrieval, we first retrieve 50
candidate memories and then select 20 memories for injection. For teacher memory
filtering, we use a minimum score threshold of 0.01. For writer-side memory
selection, we keep the top 30\% of generated memories according to the memory
score. We set the maintenance period to $Q=30$ tasks. The evolver is optimized
with the unified slow-loop objective described in \Cref{sec:slow-evolution}.

The main experiments are conducted on a Linux server with 8 NVIDIA A800 (80G) GPUs.

\section{More Results}\label{app:result}

\begin{figure}[!t]
    \centering
    \includegraphics[width=\linewidth]{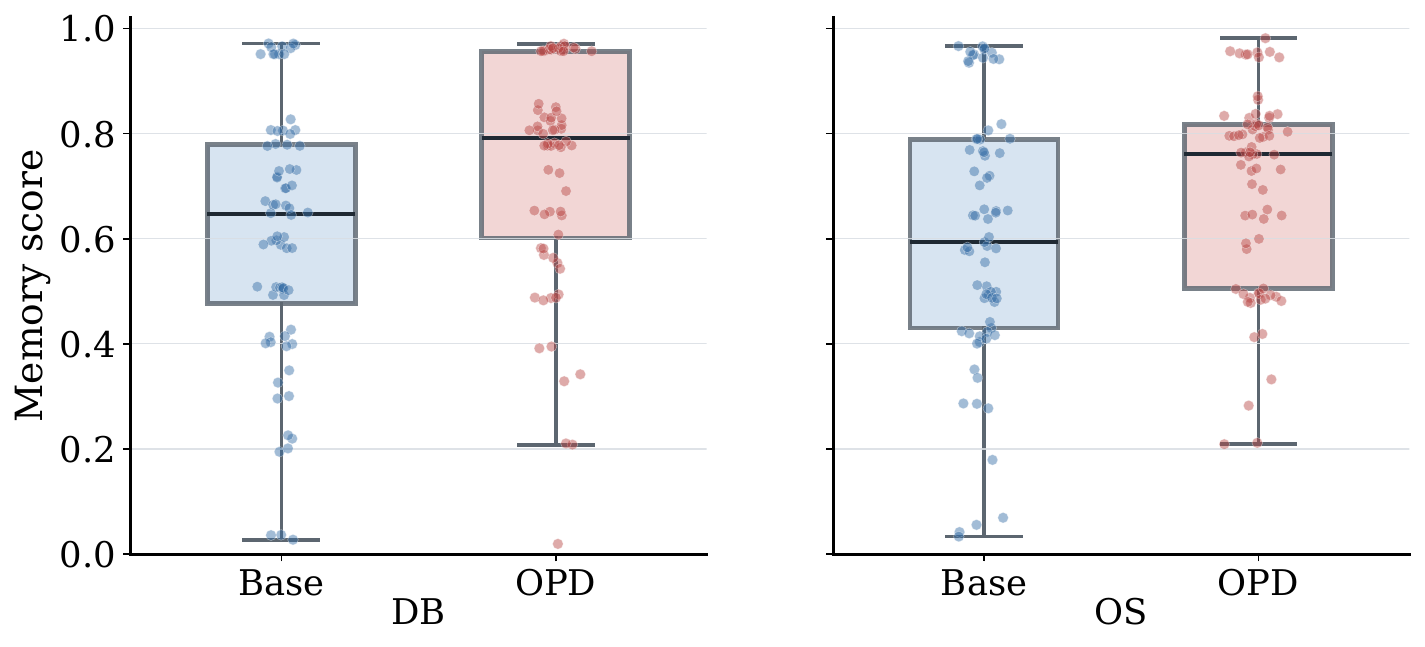}
    \caption{Additional results for memory selection on LifelongAgentBench. Left: DB; right: OS.}
    \label{fig:selection-lifelong}
\end{figure}

\begin{figure}[!t]
    \centering
    \includegraphics[width=\linewidth]{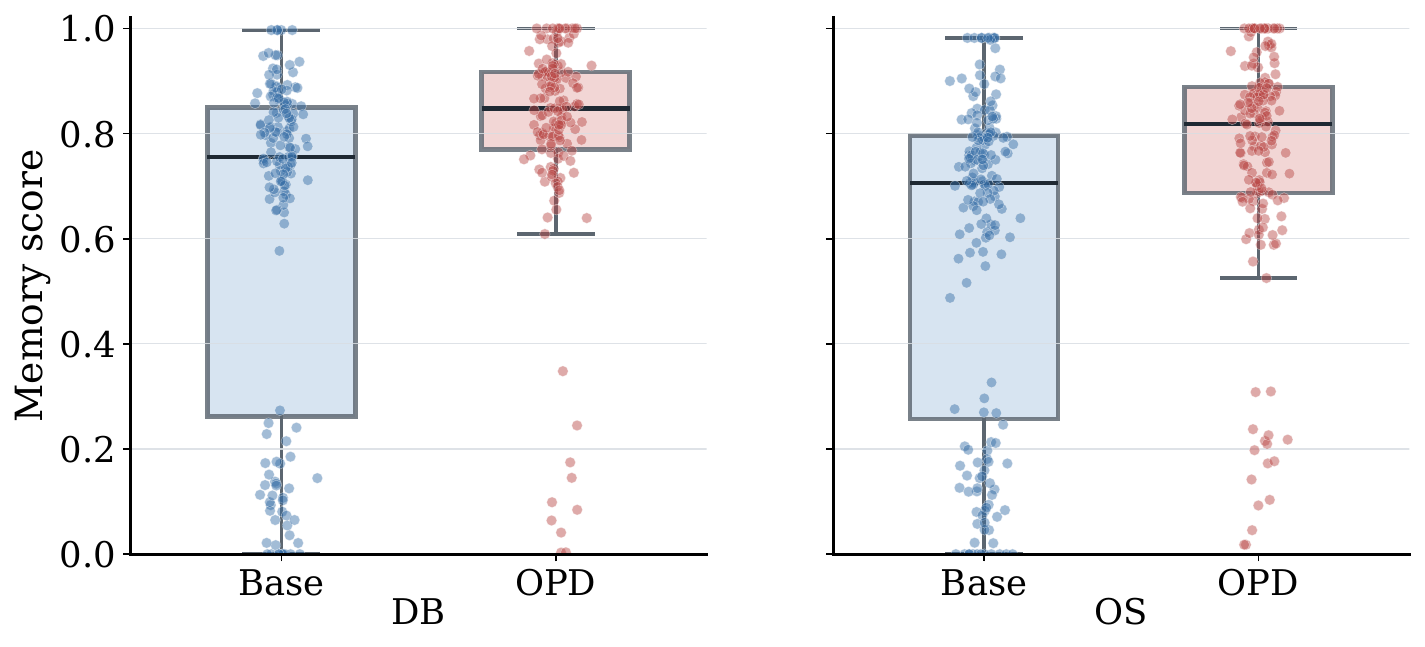}
    \caption{Additional results for memory writing on LifelongAgentBench. Left: DB; right: OS.}
    \label{fig:writing-lifelong}
\end{figure}

\Cref{fig:selection-lifelong,fig:writing-lifelong} report additional results on
LifelongAgentBench. These experiments complement the main results by evaluating
the selection and writing distillation effects on the lifelong DB and OS
environments. \ourmethod consistently shifts the memory-score
distributions upward compared with the vanilla model, indicating that slow-loop
training improves both which memories are injected and what memories are written
for future reuse.

\paragraph{Impact of Selection Distillation on LifelongAgentBench.}
\Cref{fig:selection-lifelong} compares the calibrated memory-score distributions
of memories selected by the vanilla model and by \ourmethod on LifelongAgentBench
DB and OS tasks. On both environments, \ourmethod yields higher-score selected
memories than the base model. The score distribution shifts upward, and the
low-score region is reduced, showing that the trained selector is less likely to
inject broad, noisy, or weakly related memories. This supports the claim that
selection distillation improves not only the amount of retrieved memory, but also
the task relevance of the injected memory context.

\paragraph{Impact of Writing Distillation on LifelongAgentBench.}
\Cref{fig:writing-lifelong} evaluates whether the trained writer produces
memories with higher future utility on LifelongAgentBench. Compared with the
vanilla writer, \ourmethod produces memories whose calibrated scores are higher
and more concentrated on both DB and OS tasks. The reduction of low-score
memories suggests that the trained writer is less prone to storing generic or
misleading advice, and instead writes compact memories that better capture the
causal error patterns and task-specific constraints needed for future attempts.
Together with the main results, these LifelongAgentBench experiments show that
slow-loop training generalizes beyond the primary benchmark settings and improves
both memory selection and memory writing in long-horizon agent environments.

\section{Case Study}\label{app:case}

\begin{figure}[t]
    \centering
    \includegraphics[width=0.48\linewidth]{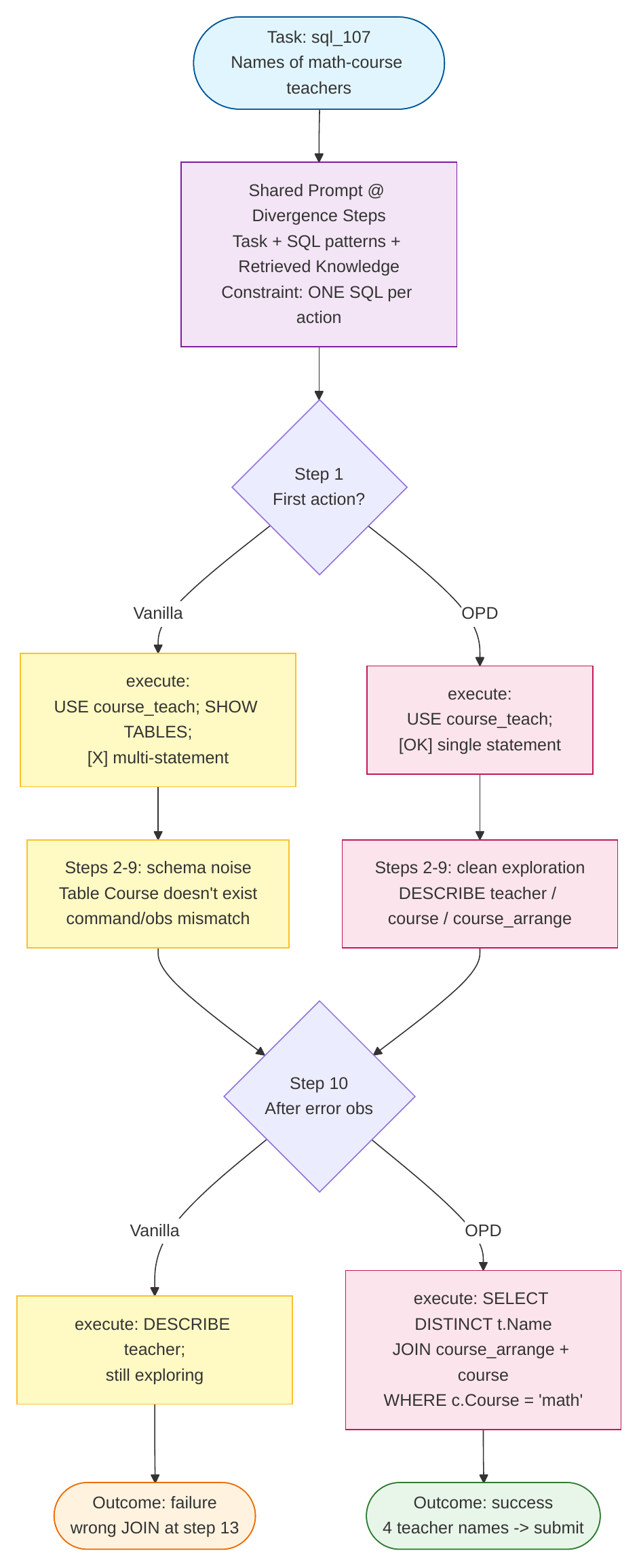}
    \hfill
    \includegraphics[width=0.48\linewidth]{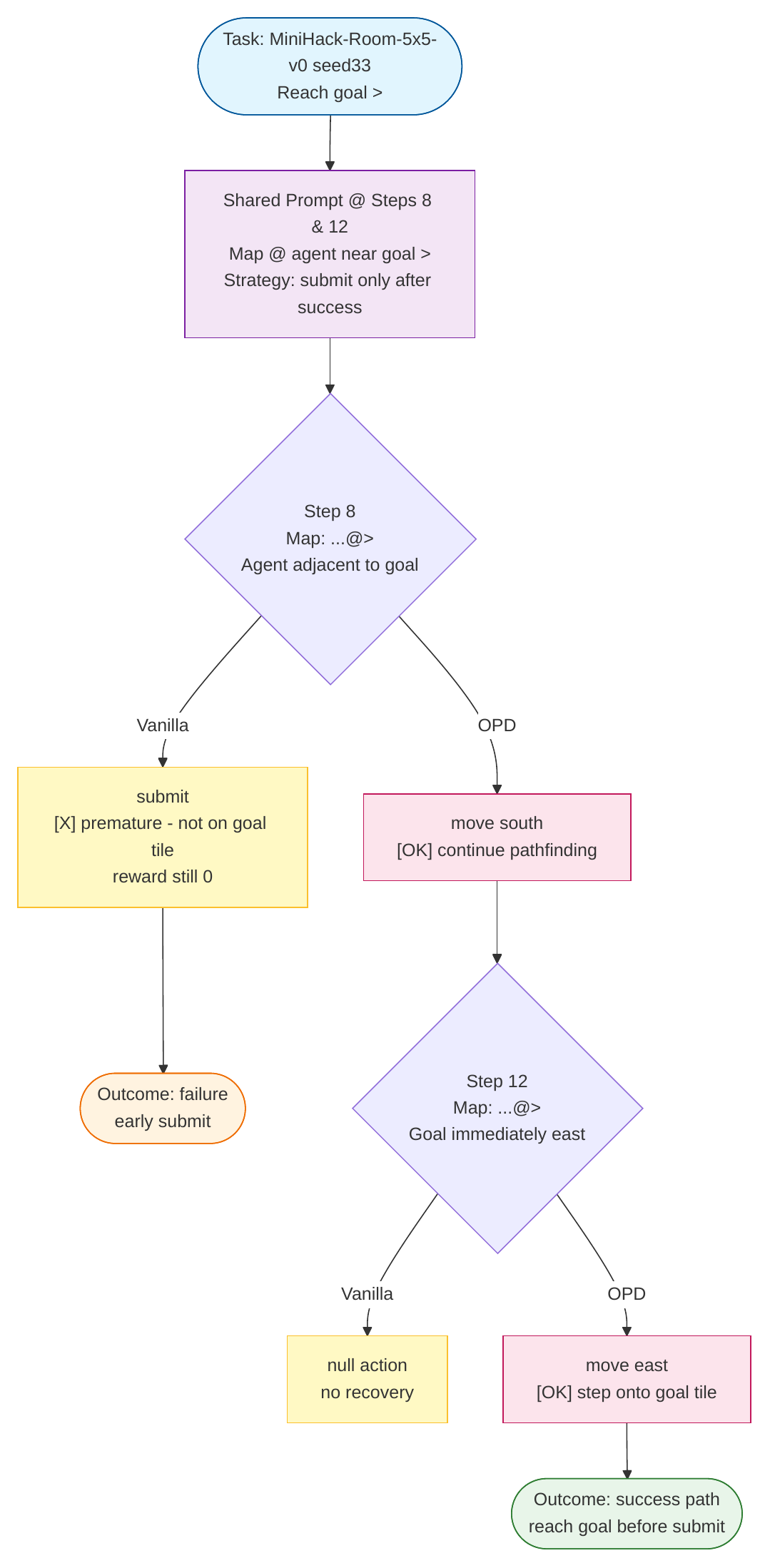}

    \vspace{0.5em}

    \includegraphics[width=0.48\linewidth]{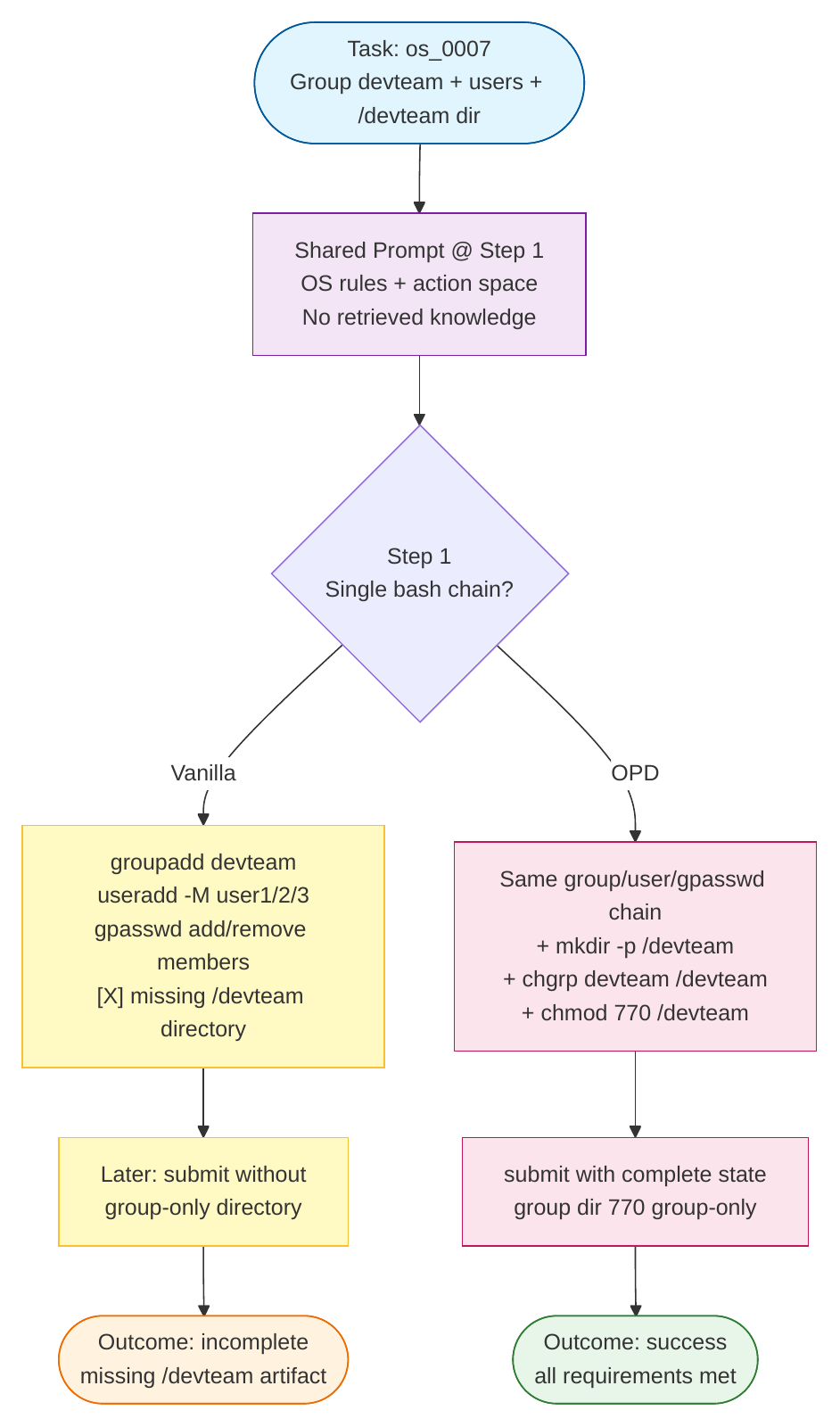}

    \caption{Case studies for the \textsc{Executor}. Top-left: InterCode-SQL; top-right: MiniHack-Room-5$\times$5; bottom: LifelongAgentBench-OS.}
    \label{fig:case-study-executor}
\end{figure}

\begin{figure}[!h]
    \centering
    \includegraphics[width=0.48\linewidth]{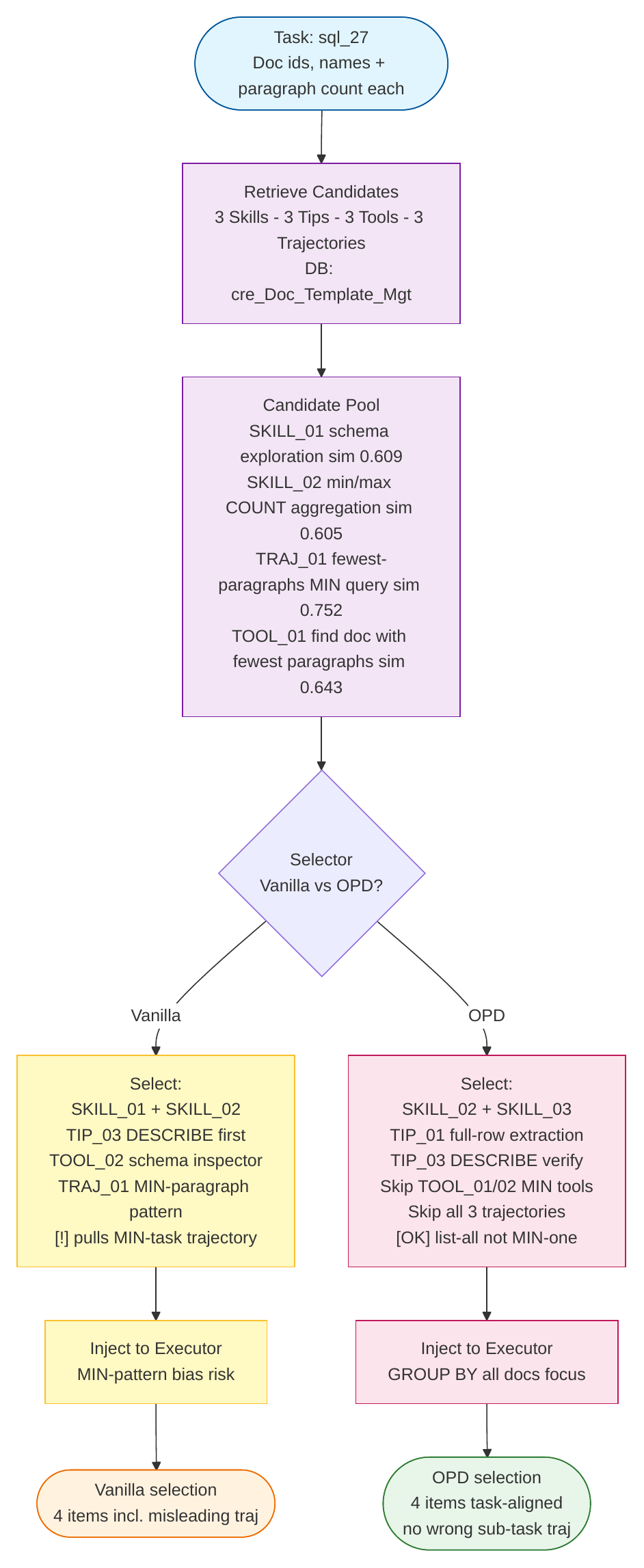}
    \hfill
    \includegraphics[width=0.48\linewidth]{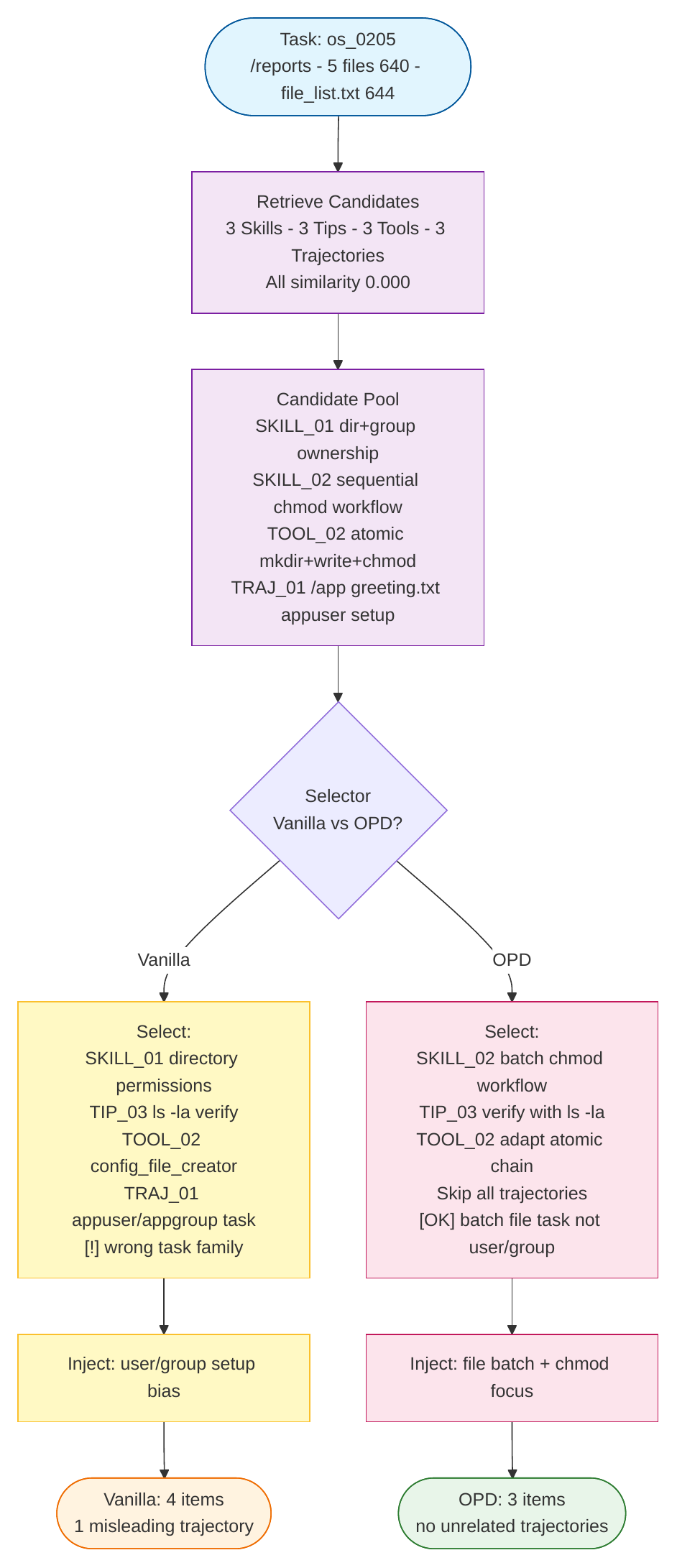}

    \vspace{0.5em}

    \includegraphics[width=0.48\linewidth]{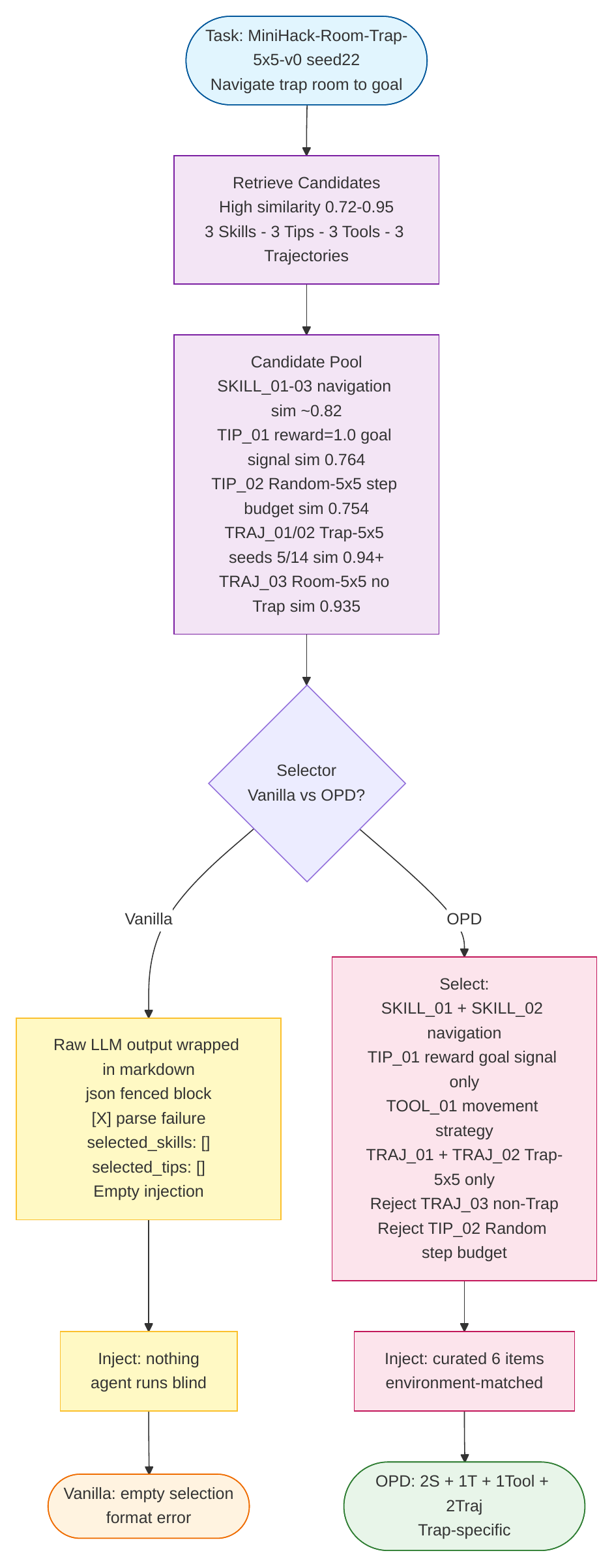}

    \caption{Case studies for the \textsc{Selector}. Top-left: InterCode-SQL; top-right: LifelongAgentBench-OS; bottom: MiniHack-Trap.}
    \label{fig:case-study-selector}
\end{figure}

\begin{figure}[!h]
    \centering
    \includegraphics[width=0.48\linewidth]{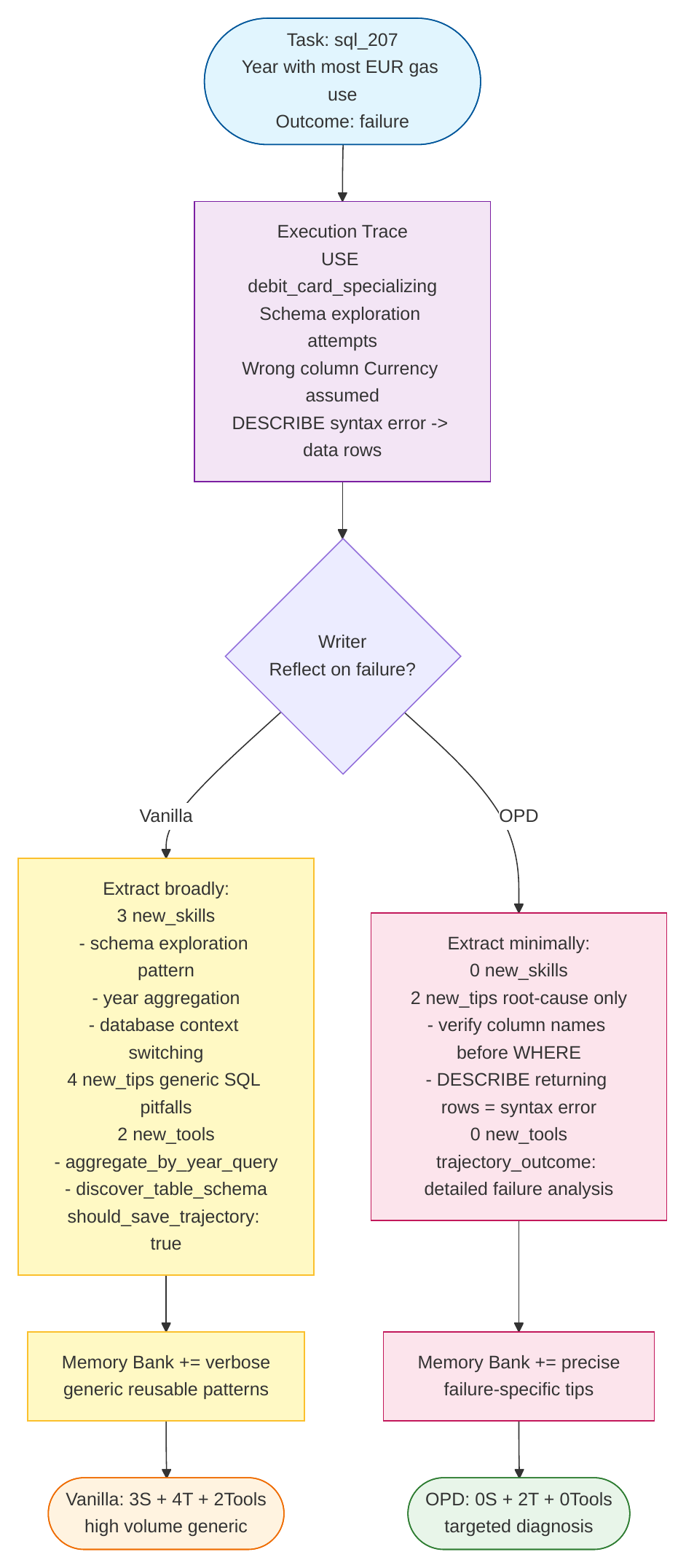}
    \hfill
    \includegraphics[width=0.48\linewidth]{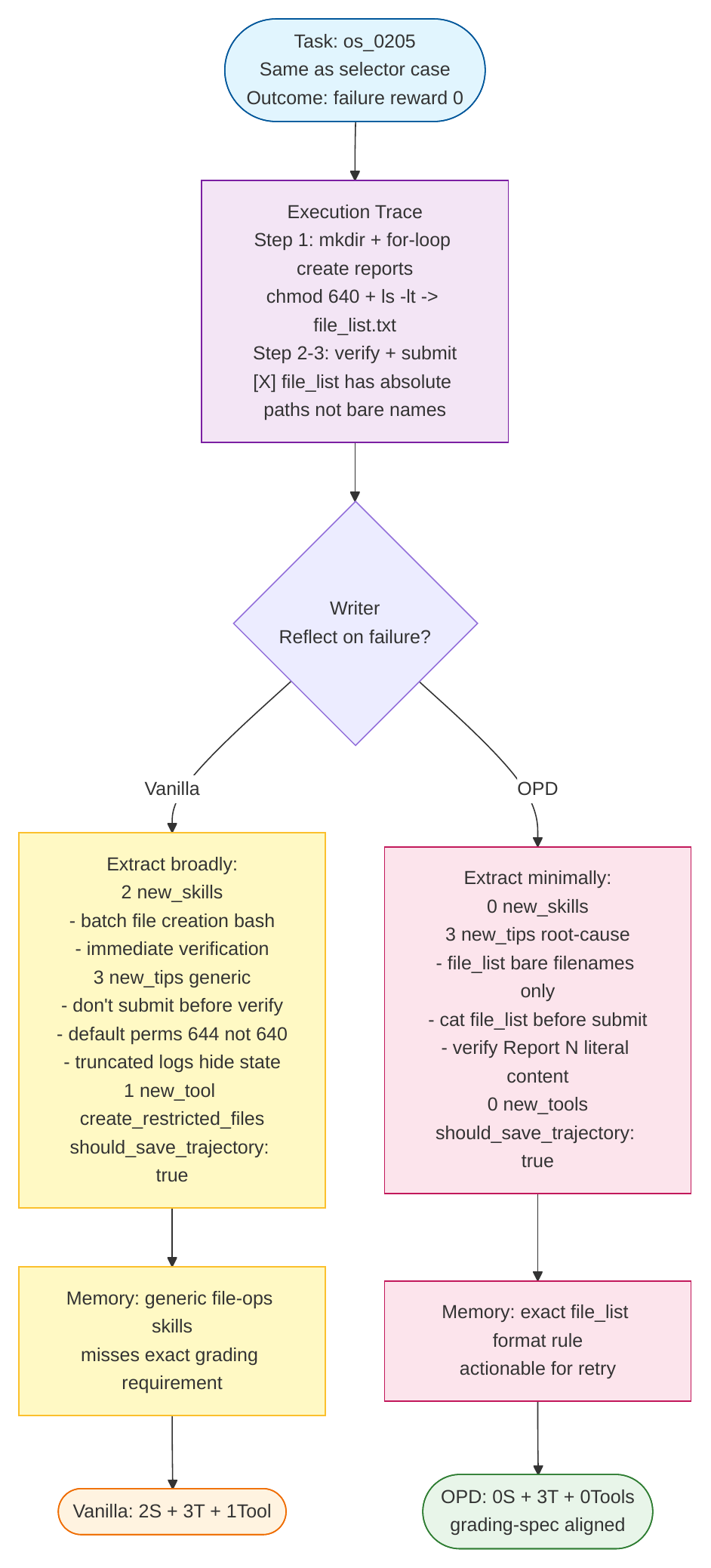}

    \vspace{0.5em}

    \includegraphics[width=0.48\linewidth]{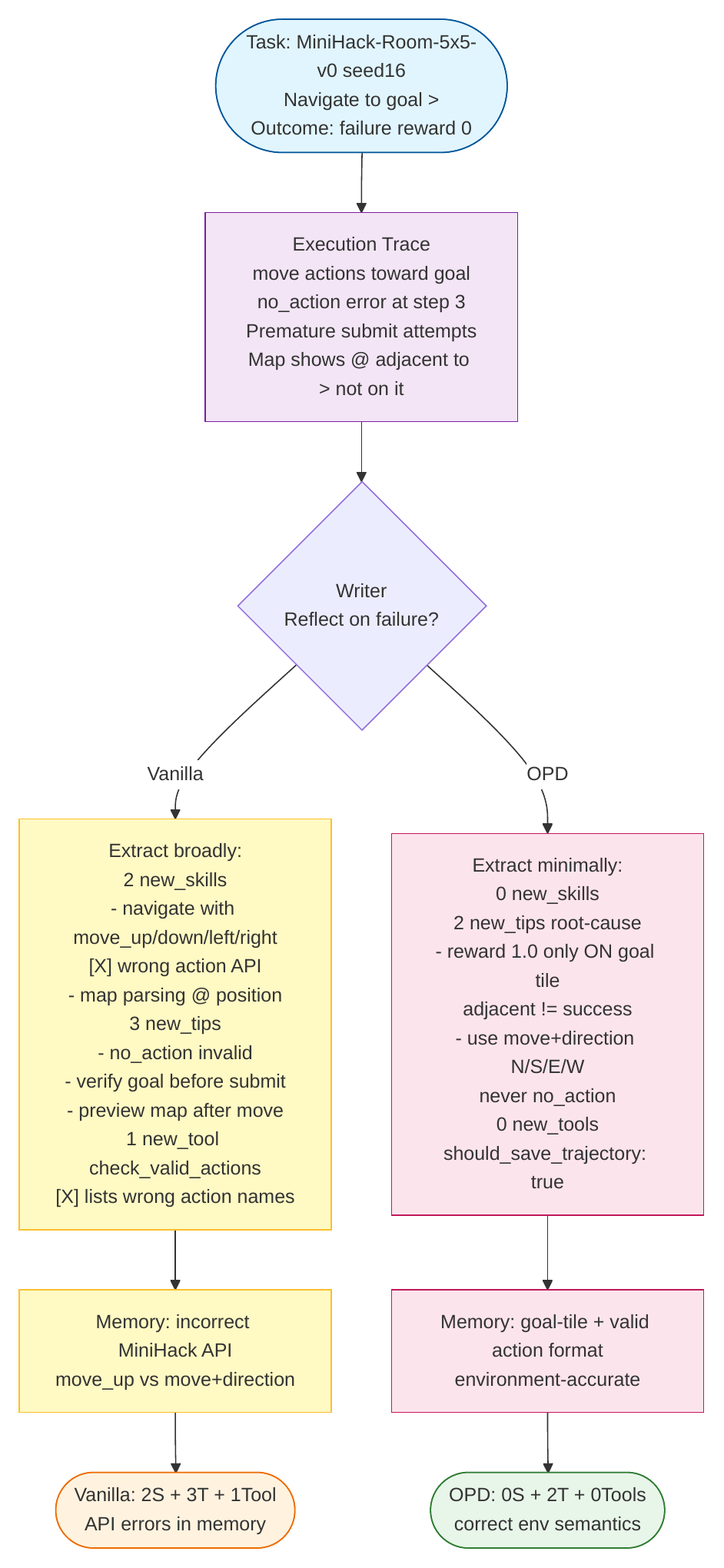}

    \caption{Case studies for the \textsc{Writer}. Top-left: InterCode-SQL; top-right: LifelongAgentBench-OS; bottom: MiniHack-Room-5$\times$5.}
    \label{fig:case-study-writer}
\end{figure}

\Cref{fig:case-study-executor,fig:case-study-selector,fig:case-study-writer}
provide qualitative examples of how slow-loop training improves the
three components of the memory evolution pipeline: execution, memory selection,
and memory writing. Across InterCode-SQL, LifelongAgentBench-OS, and MiniHack,
the vanilla models often either over-generalize from previous experience or fail
to localize the immediate cause of failure. In contrast, \ourmethod produces more
task-specific behavior: it retrieves memories that match the current task,
writes compact causal tips instead of generic advice, and executes actions that
better satisfy environment-specific constraints.

\Cref{fig:case-study-executor} focuses on the \textsc{Executor}. The examples
show that the vanilla executor can violate task-level constraints, stop before
the true success condition is reached, or omit required side effects. In
InterCode-SQL, it issues multiple statements in one action and continues schema
exploration after noisy observations. In MiniHack, it submits while merely
adjacent to the goal. In LifelongAgentBench-OS, it completes group membership
but misses the required group-owned directory. \ourmethod instead learns to
respect the operational constraints of each environment: it separates SQL
actions, continues navigation until the agent reaches the goal tile, and
performs the missing directory creation and permission-setting operations.

\Cref{fig:case-study-selector} examines the \textsc{Selector}. The vanilla
selector often injects memories that are superficially related but not causally
useful for the current task, such as choosing a minimum-count SQL trajectory for
a paragraph-counting query or a user/group setup workflow for a permission-editing
OS task. In the MiniHack-Trap example, it even produces an empty injection after
a JSON parse failure. \ourmethod filters memory more aggressively and keeps only
the skills, tips, tools, or trajectories that directly support the present
problem. This indicates that slow-loop training improves not only whether memory
is used, but also which memory is allowed to influence the next attempt.

\Cref{fig:case-study-writer} analyzes the \textsc{Writer}. The vanilla writer
tends to convert failures into broad reusable advice, and in some cases even
records incorrect environment APIs. \ourmethod instead writes narrower
future-facing memories that target the root cause of the failure. In
InterCode-SQL, it records the need to verify column names and interpret malformed
schema queries. In LifelongAgentBench-OS, it records the grading-specific
requirement to output bare filenames and verify the generated file list before
submission. In MiniHack, it records the correct action interface and the fact
that reward is obtained only by stepping onto the goal tile.

\section{Information About Use Of AI Assistants}

AI assistance was used only for writing polishing and auxiliary coding support in this work.

\end{document}